\documentclass[conf]{new-aiaa}
\usepackage[utf8]{inputenc}
\usepackage{graphicx}
\usepackage{amsmath}
\usepackage{booktabs}
\usepackage{xcolor}
\usepackage{longtable,tabularx}
\hypersetup{hidelinks}

\makeatletter\@secpenalty=0\makeatother
\newcommand{\E}{\mathbb{E}}
\newcommand{\pbar}{\bar{p}_{95}}
\newcommand{\phat}{\hat{p}}

\title{ Constraint-Aware Decision Support for\\Tactical Air Mission Planning}

\author{Hasan Taha Bagci\footnote{Researcher, Aerospace Research Center, Istanbul Technical University, Istanbul, Turkiye, 34485; bagcih21@itu.edu.tr (Corresponding Author)}}
\affil{Aerospace Research Center, Istanbul Technical University, Istanbul, 34485, Türkiye}
\author{Fatma Elif Yalcindag\footnote{Department of Electronics and Communication Engineering, Istanbul Technical University, Istanbul, Turkiye, 34469; yalcindag22@itu.edu.tr}}
\affil{Department of Electronics and Communication Engineering, Istanbul Technical University, Istanbul, 34469, Turkiye}
\author{Emre Koyuncu\footnote{Associate Professor, Aerospace Research Center, Istanbul Technical University, Istanbul, Turkiye, 34485; emre.koyuncu@itu.edu.tr}}
\affil{Aerospace Research Center, Istanbul Technical University, Istanbul, 34485, Türkiye}
\author{Nazim Kemal Ure\footnote{Senior Research Advisor, Stanford Intelligent Systems Laboratory, Stanford University, Stanford, California, 94305; ure@stanford.edu}}
\affil{Stanford Intelligent Systems Laboratory, Stanford University, Stanford, California, 94305}

\begin{document}

\maketitle

\begin{abstract}
Planning an air strike means choosing a course of action (COA) under limits that are not negotiable: crewed aircraft must survive, fuel reserves must hold, and closed airspace must not be entered. This paper presents CONCORD, a framework in which a learned policy generates a complete COA in a single decision, with standing down as an explicit option, and in which the word ``constraint'' has one of three precise meanings: a limit that can be checked before the mission is flown, a limit on the violation rate averaged over scenarios, or a limit on the violation rate in every scenario. For the limit in every scenario we propose HardNet-PPO, in which the HardNet enforcement layer confines a proximal policy optimization (PPO) policy to a certified set, the COAs that a learned and calibrated risk model admits. We compare it with unconstrained PPO and five constrained baselines in the Command: Modern Operations (CMO) wargame simulator, each with the same budget of 8,000 training missions, on five scenarios and three seeds. Every trained policy is tested on 200 held-out draws per scenario, for the COA it recommends and for COAs sampled from it, with a one-sided 95\% Clopper--Pearson bound on every violation rate against a level of 5\%. Unconstrained PPO earns the highest reward but violates a constraint in 44\% of its test missions, and a baseline that constrains the average can meet its budget and still violate almost four times too often in a single scenario. HardNet-PPO is the only learner whose estimated violation rate stays within the level in every case, for recommended and sampled COAs alike. It also incurs the fewest violations during training, 4\% of its missions, and its reward is on par with the best constrained baseline.
\end{abstract}

\section{Nomenclature}

\vspace{2pt}
{\renewcommand\arraystretch{1.0}
\noindent\begin{tabular}{@{}l@{\;\,=\;\,}p{5.55cm}@{\hspace{0.6cm}}l@{\;\,=\;\,}p{5.55cm}@{}}
$\mathcal{A}$, $a$ & set of COAs; a COA & $\phat$, $\pbar$ & estimated violation rate; its upper bound \\
$\mathcal{A}_K(s)$ & COAs that pass the known constraints & $R$ & reward of a mission \\
$\mathcal{A}_\tau(s)$ & certified set of COAs & $\mathcal{S}$, $s$ & set of mission contexts; a context \\
$a_0$ & stand-down COA & $\mathcal{Z}$, $z$ & set of scenarios; a scenario \\
$C$ & cost of a mission, 1 if a limit is violated & $\delta$ & violation level in each scenario \\
$c(s,a)$ & violation probability of $a$ in context $s$ & $\eta$ & commit threshold \\
$\hat{c}_j(s,a)$ & prediction of $c(s,a)$ by classifier $j$ & $\theta$ & policy parameters \\
$d$ & cost budget for the expected cost & $\lambda$ & Lagrange multiplier \\
$J_R$, $J_C$ & expected reward; expected cost & $\mu$, $\mu_z$ & distribution of contexts; in scenario $z$ \\
$J_C^z$ & expected cost in scenario $z$ & $\pi$ & policy \\
$k$, $n$ & violations and missions of a test & $\tau_z$ & certificate threshold of scenario $z$ \\
\end{tabular}}

\section{Introduction}

\lettrine{P}{lanning} an air strike means deciding how to suppress the air defenses, which routes to fly, which weapons to release, and under what condition to press the attack. These decisions are taken under limits that are not negotiable: crewed aircraft return, each with a fuel reserve, and closed airspace is not entered. The package can always stand down. A tool that proposes a course of action (COA) to a planning staff under time pressure is useful only if it can state how often its proposal would break such a limit, and if it recommends standing down only when no acceptable COA exists. Simulation-based decision support for this task has long been proposed \cite{surdu2008deepgreen}, and learning-based methods now generate plans \cite{goecks2024coagpt} or control platforms \cite{pope2023hierarchical}. To our knowledge, however, none of them attaches a tested violation probability to the plan it returns.

\begin{figure}[hbt!]
\centering
\includegraphics[width=\textwidth]{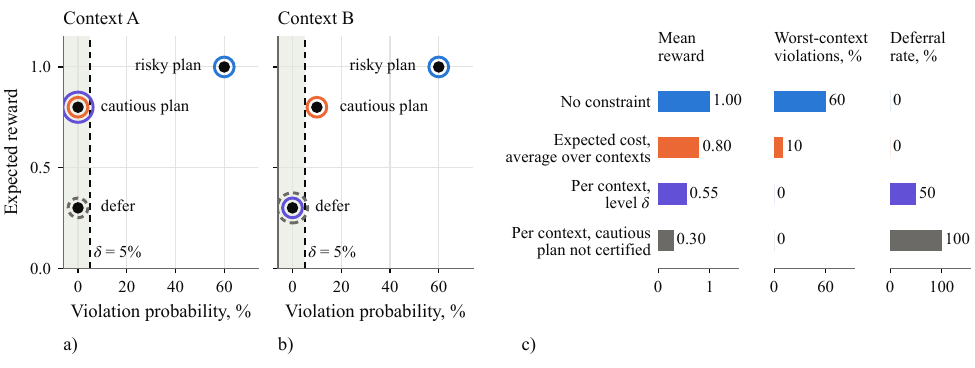}
\caption{A decision example with two equally likely contexts and three options. a), b) Expected reward and violation probability of the options in contexts A and B; a ring marks the option selected under each reading of the instruction, in the colors of c), and the dashed ring marks the selection when the cautious plan cannot be certified; the shaded band is the admissible region at the level $\delta = 5\%$. c) Mean reward, violation rate in the worst context, and deferral rate that result.}
\label{fig:example}
\end{figure}

Stating ``how often'' hides a choice. Suppose that a violation is tolerable in at most 5\% of the decisions, and consider a planner who faces one decision in each of two equally likely contexts, A and B (Fig.~\ref{fig:example}). Three options exist in both. A risky plan earns an expected reward of 1.0 and violates a limit in 60\% of its executions. A cautious plan earns 0.8; it never violates in context A and violates in 10\% of its executions in context B. Deferring the decision earns 0.3 and violates nothing. If the 5\% is ignored, the risky plan is chosen everywhere. If the 5\% bounds the violation rate averaged over contexts, which is the expected-cost constraint of a constrained Markov decision process (CMDP), the cautious plan is optimal in both contexts: the average rate is exactly 5\% and the reward is 0.80, but every violation falls in context B, at a rate of 10\%. If the 5\% must hold for the plan chosen in each context, the cautious plan is excluded in B, where only deferral remains; the mean reward falls to 0.55, and half of the decisions are deferred.

A fourth outcome matters most in practice. A learner is not given the violation probabilities; it has to estimate them from trials. Showing with 95\% confidence that the cautious plan violates in at most 5\% of its executions in context A takes at least 59 trials without a violation, and a learner that is penalized or blocked whenever it approaches a risky option may never collect them. Such a learner defers in both contexts, earns 0.30, and shows a perfect safety record. Deferral always satisfies the constraint and reveals nothing: from the violation record alone, a learner that failed to find an admissible plan cannot be distinguished from one for which none exists.

Constrained reinforcement learning (RL) offers methods for each reading, but these methods are compared almost exclusively on continuous-control benchmarks and under expected-cost constraints \cite{ray2019benchmarking,ji2023safetygymnasium}, the weaker of the two readings. This paper brings the comparison to air strike planning. We present CONCORD, a constraint-aware decision-support framework in which a policy generates a complete COA from the mission context. The policy composes the timing of the defense suppression, the routes, the weapon employment, the use of decoys, and a commit rule, instead of ranking candidates that a planner has prepared, and standing down is one of the COAs that it can generate. The paper makes three contributions. It formulates mission-level COA generation as a single-decision constrained problem that keeps apart constraints known before execution, an expected-cost constraint over scenarios, and a per-scenario constraint (Sec.~\ref{sec:formulation}). It proposes HardNet-PPO, a new certified-set learner that places PPO \cite{schulman2017ppo} behind the HardNet enforcement layer \cite{min2024hardnet} and confines it to the COAs admitted by a learned and calibrated risk model, with the violations of its exploration held to a ration (Sec.~\ref{sec:hardnet}). And it compares this learner with unconstrained PPO and five constrained baselines under one budget of simulated missions in the high-fidelity wargame simulator Command: Modern Operations (CMO), testing every trained policy scenario by scenario on held-out draws with a Clopper--Pearson bound on each violation count \cite{clopper1934}. The comparison shows averages that hide one bad scenario, a certified-set policy that stays within the level in every scenario, and an exploration problem that no learner solves reliably (Secs.~\ref{sec:setup}--\ref{sec:discussion}).

\section{Related Work}
\label{sec:related}

\subsection{Course-of-Action Decision Support and Learning for Mission Planning}

Simulation-based support for COA development predates learning methods. The Deep Green concept proposed simulating the possible futures of an ongoing operation and alerting the commander when the plan departs from them \cite{surdu2008deepgreen}. Language-model approaches such as COA-GPT generate candidate COAs and revise them with human feedback \cite{goecks2024coagpt}; they return a plan quickly, without a tested probability that the plan breaks a stated limit.

RL has been applied to air mission planning in purpose-built simulators. Zhang et al.\ trained PPO agents to plan missions of strike aircraft, jammers, and decoys against surface-to-air missile sites \cite{zhang2020airdominance}, and the planning of a suppression mission for unmanned aircraft has been cast as a Markov decision process and solved with PPO \cite{yue2022uavmission}. At the platform level, hierarchical RL agents were developed for simulated air combat at the AlphaDogfight Trials \cite{pope2023hierarchical}. A systematic review covers RL for decision support in defense and security \cite{schadd2026review}, and artificial intelligence and deep RL in wargaming are treated in Refs.~\cite{goodman2020wargaming} and~\cite{delrio2026wargaming}. To our knowledge, none of these studies states operational limits as constraints with a violation level that is tested on held-out draws.

CMO and simulators of the same family have served for Monte Carlo analysis of an operational vignette with 100 iterations per design point \cite{hagley2021commandpe}, as the environment of a PPO agent that controls units in custom scenarios \cite{dimitriu2024cmo}, and as a test bed in which a language model advises a human player \cite{chen2024llmwargaming}. These three works use purpose-built scenarios, as we do, and to our knowledge none of them treats rules of engagement or loss limits as explicit constraints.

\subsection{Constrained Reinforcement Learning}

The CMDP adds to a Markov decision process one or more cost functions with limits on their expected return \cite{altman1999cmdp}. Constrained policy optimization (CPO) enforces the limits approximately at every policy update inside a trust region \cite{achiam2017cpo}, Lagrangian methods adapt a penalty multiplier \cite{ray2019benchmarking,tessler2019rcpo}, and first-order constrained optimization in policy space (FOCOPS) solves the constrained update in the nonparameterized policy space and projects the solution back onto the parametric policy \cite{zhang2020focops}. All of them constrain an expectation.

Surveys of constraint formulations \cite{wachi2024survey} and of state-wise safe RL \cite{zhao2023statewise} separate the expected-cost constraint from chance, state-wise, and almost-sure constraints. Absolute state-wise constrained policy optimization (ASCPO) constrains an upper probability bound of the maximum state-wise cost, its mean plus a multiple of its variance, instead of the mean alone \cite{zhao2024ascpo}; chance-constrained policy optimization tightens the constraints with backoffs \cite{petsagkourakis2022chance}; and augmenting the state with the remaining safety budget yields almost-sure satisfaction \cite{sootla2022saute}.

A second family restricts the action set and leaves the update rule unchanged. Shields replace actions that would violate a safety specification \cite{alshiekh2018shielding}, safety layers correct each action with a learned linear cost model \cite{dalal2018safetylayer}, safety critics and recovery policies veto or override actions whose estimated risk exceeds a threshold \cite{srinivasan2020safetycritic,thananjeyan2021recovery}, and invalid-action masking removes inadmissible actions from a categorical policy \cite{huang2022masking}. HardNet appends a differentiable closed-form layer that makes the output of a network satisfy input-dependent affine constraints by construction \cite{min2024hardnet}. In bandits with safety constraints, the single-decision counterpart of our problem, learning starts from actions known to be safe and expands the safe set as its estimate sharpens \cite{amani2019safebandits}. The certified-set learner of Sec.~\ref{sec:methods} combines these elements: a known safe action, a learned risk model, an enforcement layer, and a calibration step.

\subsection{Benchmarks and Evaluation Practice}

Safe RL methods are compared predominantly on continuous-control suites \cite{ray2019benchmarking,ji2023safetygymnasium}. SafeOR-Gym moves the comparison to operations research problems with structured and mixed-integer decisions and shows training reward and cost side by side, with evaluation reported separately. It finds on-policy trust-region methods the most stable, with the Lagrangian variant of trust-region policy optimization the strongest overall \cite{ramanujam2026safeorgym}. For neural routing solvers, Bi et al.\ show that feasibility masking fails when feasibility cannot be decided step by step and combine a Lagrangian term with a preventative mask \cite{bi2024constraints}. Our setting differs from both: an episode is one decision, its outcome is stochastic, and it is produced by a high-fidelity simulator at a cost of seconds per episode.

A simulator of this cost permits few training runs. Agarwal et al.\ show that point estimates from a handful of runs are unreliable and recommend interval estimates \cite{agarwal2021precipice}. We therefore report reward and deferral with their range over the training seeds, and we compute the bound on a violation rate for each trained policy separately instead of pooling policies.

\section{Problem Formulation}
\label{sec:formulation}

\subsection{Course-of-Action Generation as a Single Decision}

\begin{figure}[hbt!]
\centering
\includegraphics[width=\textwidth]{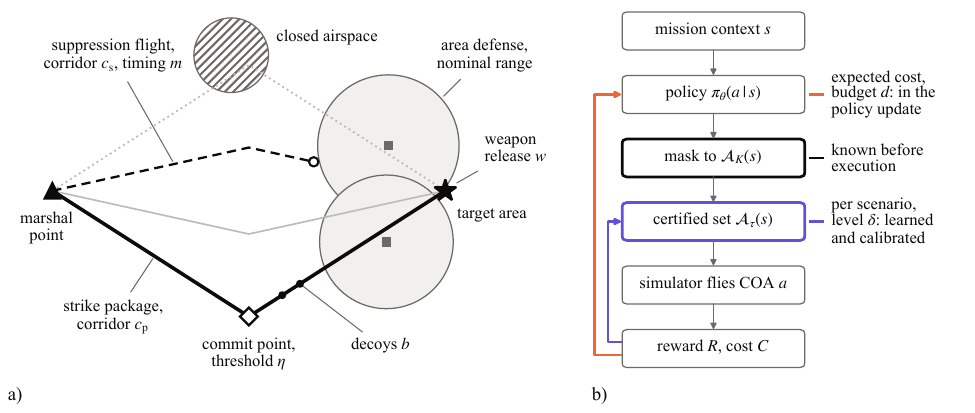}
\caption{Mission-level COA generation as a single decision. a) Elements of a mission, with the six factors of a COA, Eq.~\eqref{eq:coa}, marked where they act; four of the $N_c$ corridors are drawn, and the dotted one crosses closed airspace. The layout is illustrative and is not one of the scenarios. b) Decision loop and the place at which each notion of constraint enters: known constraints mask the action set, the per-scenario level restricts it further to a certified set, and the expected-cost budget enters the policy update.}
\label{fig:problem}
\end{figure}

Figure~\ref{fig:problem}a shows the elements of a mission. A strike package of crewed aircraft waits at a marshal point, from which dog-leg corridors lead to a target area protected by area-defense sites. A suppression flight can attack these sites, and expendable decoys can fly ahead of the package. At the dog-leg of its corridor, the commit point, the package either presses on to the target or turns back.

A mission context $s \in \mathcal{S}$ describes what the planner knows before the mission: the friendly package, the believed positions and classes of the defensive systems, the targets, the closed airspace, and the constraint thresholds. The policy receives it as a vector of 304 numbers in coordinates aligned with the direct track from the marshal point to the targets. The vector has a fixed layout in which unused slots are zero: 16 numbers for each of up to ten aircraft (position, fuel, role, and weapon loads with their ranges), 10 for each of up to eight defensive systems (believed position, engagement range, class, and the uncertainty of the position), 7 for each of up to six targets (position, priority, hardness, and the share of the close-in and of the standoff load that the target would take), 4 for each of up to two circles of closed airspace, and 14 scalars (the constraint thresholds, the mission time limit, the size of the package, and the corridor geometry, including the position of a border). Contexts are drawn from $\mu = |\mathcal{Z}|^{-1}\sum_{z\in\mathcal{Z}}\mu_z$, where each scenario $z$ fixes the terrain, the package, and the composition of the defense, and $\mu_z$ perturbs the layout from draw to draw. A policy $\pi(a\,|\,s)$ generates one COA
\begin{equation}
\label{eq:coa}
a = (m,\; c_{\mathrm{p}},\; c_{\mathrm{s}},\; w,\; b,\; \eta) \in \mathcal{A},
\end{equation}
where $m \in \{\text{first}, \text{simultaneous}, \text{none}\}$ is the timing of the suppression flight relative to the strike package, $c_{\mathrm{p}}, c_{\mathrm{s}} \in \{0,\ldots,N_c-1\}$ are the corridors of the package and of the suppression flight, $w \in \{\text{close-in}, \text{standoff}\}$ is the weapon-employment mode, $b \in \{0,1\}$ sends the decoys ahead of the package, and $\eta \in \{0, 0.25, 0.5, 0.75, 1\}$ is the commit threshold: the package proceeds beyond its commit point only if the surviving fraction of the area-defense sites is at most $\eta$. With $N_c = 12$ corridors, the set $\mathcal{A}$ contains $3\times12\times12\times2\times2\times5 = 8{,}640$ COAs. The policy composes a COA factor by factor; no list of candidate COAs is prepared for it to rank.

A scripted executor flies the generated COA in the simulator, and the episode ends when the package has recovered or the mission time has elapsed. The episode returns a reward $R$ and a cost $C$. An episode is thus one mission; we write ``mission'' for an episode of our problem and keep ``episode'' for the general setting of the published methods. Because an episode contains one decision, the CMDP reduces to a constrained contextual bandit: there is no discounting and no bootstrapping, and the value of a COA is the expectation of its outcome. We keep the CMDP terminology of reward, cost, cost budget, and policy \cite{achiam2017cpo,ramanujam2026safeorgym}, and we write $c(s,a) = \Pr[C = 1 \,|\, s,a]$ for the violation probability of a COA in a context.

\subsection{Reward and Cost}

The reward is a weighted scorecard of the mission outcome,
\begin{equation}
\label{eq:reward}
R = 0.40\,S + 0.35\,V - 0.10\,E - 0.08\,\min(U/u^\star, 1) - 0.05\,F + 0.02\,T ,
\end{equation}
where $S$ is the priority-weighted fraction of the target value destroyed, $V$ the fraction of crewed aircraft that survive, $E$ the fraction of the mission time that crewed aircraft spend inside the nominal range of a live defense, $U$ the number of expendable aircraft lost, with budget $u^\star$, $F$ the mean fraction of fuel used by the surviving crewed aircraft, and $T$ an indicator that weapons are released within the mission time window. The destruction of the targets and the survival of the crewed aircraft are the primary objectives. The weights are a modeling choice and were fixed before the experiments.

Three hard constraints define the cost,
\begin{equation}
\label{eq:cost}
C = \mathbf{1}\bigl[\,V < 0.90 \;\lor\; f_{\min} < 0.15 \;\lor\; t_{\mathrm{RoE}} > 0\,\bigr] ,
\end{equation}
where $f_{\min}$ is the smallest fuel fraction at recovery among the surviving crewed aircraft and $t_{\mathrm{RoE}}$ the time that crewed aircraft spend in closed airspace, which the rules of engagement (RoE) forbid. A mission violates when the surviving fraction of crewed aircraft falls below 0.90, when a surviving crewed aircraft recovers with less than 15\% of its fuel, or when a crewed aircraft enters closed airspace. Each scenario fields four or six crewed aircraft, strike package and suppression flight together, so the loss of a single crewed aircraft is a violation. Crewed survival appears in both Eq.~\eqref{eq:reward} and Eq.~\eqref{eq:cost}. The fuel floor and the closed airspace appear in the cost only, so an unconstrained learner has no signal for them other than their side effects on the reward.

\subsection{Three Notions of Constraint}

The word ``constraint'' carries three meanings in this problem, and each calls for a different mechanism (Fig.~\ref{fig:problem}b). The first is the expected-cost constraint of the CMDP. With $J_R(\pi) = \E_{s\sim\mu,\,a\sim\pi}[R]$ and $J_C(\pi) = \E_{s\sim\mu,\,a\sim\pi}[C]$, the problem is
\begin{equation}
\label{eq:cmdp}
\max_{\pi}\; J_R(\pi) \quad \text{subject to} \quad J_C(\pi) \le d ,
\end{equation}
in which the cost budget $d$ bounds the violation rate averaged over all contexts. A policy can satisfy Eq.~\eqref{eq:cmdp} by never violating in the easy scenarios and violating often in a hard one, as the cautious plan does in the introductory example.

The second is a per-scenario constraint, which removes the average over scenarios:
\begin{equation}
\label{eq:perscenario}
J_C^z(\pi) = \E_{s\sim\mu_z,\,a\sim\pi}[C] \le \delta \quad \text{for every } z \in \mathcal{Z} .
\end{equation}
Because $C$ is an indicator, $J_C$ and $J_C^z$ are violation probabilities; the two constraints differ in the distribution over which the probability is taken. Because survival and fuel are random outcomes, the level $\delta$ is a probability and not ``never.'' A sufficient condition for Eq.~\eqref{eq:perscenario} is that every COA the policy can generate is safe in its own context,
\begin{equation}
\label{eq:perdecision}
c(s,a) \le \delta \quad \text{for every } s \text{ and every } a \text{ with } \pi(a\,|\,s) > 0 .
\end{equation}
With $d = \delta$ the three conditions are nested: Eq.~\eqref{eq:perdecision} implies Eq.~\eqref{eq:perscenario}, which implies the constraint of Eq.~\eqref{eq:cmdp}. We set $d = \delta = 0.05$ for every constrained learner, so that the learners differ in the kind of constraint and not in its level. In the introductory example a context plays the role of a scenario and each option that of one COA, and the reading ``for the plan chosen in each context'' is the per-decision condition of Eq.~\eqref{eq:perdecision}. Under Eq.~\eqref{eq:perscenario} alone, a policy that mixes the cautious plan and deferral equally in context B would also be admissible, with a mean reward of 0.675.

The third meaning needs no learning. Whether the planned route of the strike corridor crosses closed airspace can be computed from the context before the mission is flown. Such known constraints define the set $\mathcal{A}_K(s) \subseteq \mathcal{A}$ of COAs that pass them, and a policy can be confined to $\mathcal{A}_K(s)$ by masking \cite{huang2022masking}. The mask covers the strike corridor only; whether the suppression flight strays into closed airspace depends on how the mission unfolds and has to be learned from the cost.

\subsection{Deferral}

A COA with $m = \text{none}$ and $\eta < 1$ can never meet its commit condition, because nothing attacks the area defenses; the executor does not launch the package. Such a stand-down COA violates nothing by construction and earns only the survival term of Eq.~\eqref{eq:reward} less a small fuel term, $R \approx 0.34$. Of the 8,640 COAs, 2,304 are this same decision, with rewards that differ by at most 0.004 through the weapon load; $a_0$ denotes one fixed representative, so that $c(s,a_0) = 0$. A package that launches and turns back at the commit point because its condition is not met has also not committed to the attack. We call both cases a deferral and report the deferral rate, the fraction of missions in which the package does not commit; a package that is lost before it reaches its commit point also counts as not committed. The two cases differ in risk: a mission that launches and turns back can still violate, because the suppression flight is crewed and the route to the commit point is flown. Standing down plays the role of ``defer'' in Fig.~\ref{fig:example}: it is the known safe action from which safe exploration can start, and the outcome into which an overcautious learner collapses.

\section{Methods}
\label{sec:methods}

\subsection{One Policy, One Training Loop}

Figure~\ref{fig:problem}b shows how the parts of the framework are connected. In every epoch the environment draws 200 mission contexts, 40 from each scenario. For each context the policy generates one COA, the executor flies it in the simulator (Sec.~\ref{sec:executor}), and the mission returns a reward $R$ and a cost $C$. After the 200 missions the learner updates its networks, and the next epoch begins. The seven learners share all of this, including the simulator interface, the networks, and the logging. They differ only in the update rule and in the set of COAs from which the policy may choose, so that differences in the results come from the handling of the constraint.

The policy, a reward critic $\hat{R}(s)$, and a cost critic $\hat{C}(s)$ are separate multilayer perceptrons with two hidden layers of 64 tanh units, as in the reference implementation of Ray et al.\ \cite{ray2019benchmarking}. The policy has one categorical output for each of the six factors of Eq.~\eqref{eq:coa} on a shared trunk, and the probability of a COA is the product of the six. A choice that is unavailable in a context receives zero probability through a mask on its logit \cite{huang2022masking}; the known constraints of Sec.~\ref{sec:formulation} use the same mechanism. Because a mission is one decision, the critics simply predict the expected reward and cost of a context, and the advantages of a flown COA are $A_R = R - \hat{R}(s)$ and $A_C = C - \hat{C}(s)$; neither discounting nor generalized advantage estimation applies. One policy is trained on all five scenarios and sees the scenario only through the context.

All learners are implemented in one code base. Every published method keeps its update rule, its losses, and its default hyperparameters and was checked against its authors' code. Table~\ref{tab:methods} lists the learners with every departure that the single-decision, discrete problem forces, and Appendix~A lists the settings. Before a learner was run in the simulator, it had to solve a single-context version of the introductory example within the same budget of missions and to train on a low-fidelity surrogate of the simulator.

\begin{table}[hbt!]
\caption{Learners, the constraint each one targets, and the departures from the published method that the single-decision, discrete problem forces. Departures common to all learners: a product of six categorical distributions as the policy; masking of unavailable choices; one-decision episodes without discounting; epochs of 200 missions.}
\label{tab:methods}
\centering
\footnotesize
\begin{tabular}{@{}p{1.75cm}>{\raggedright\arraybackslash}p{3.3cm}p{10.4cm}@{}}
\toprule
Learner & Constraint targeted & Reference and method-specific departures \\
\midrule
PPO & none & PPO \cite{schulman2017ppo} as implemented in Ref.~\cite{ray2019benchmarking}; no departure beyond the common ones. \\
PPO-Lag & expected cost, Eq.~\eqref{eq:cmdp} & PPO-Lagrangian \cite{ray2019benchmarking}; the cost budget $d$ applies to a one-decision episode instead of a sum of costs over a long episode. \\
PPO-Lag+M & expected cost, with $\mathcal{A}_K$ & PPO-Lag with a mask (M) on the strike corridors whose planned route crosses closed airspace; it receives the same prior knowledge as HardNet-PPO. \\
CPO & expected cost & CPO \cite{achiam2017cpo}; cost budget $d$; no rescaling of the cost surrogate by the episode length. \\
FOCOPS & expected cost & FOCOPS \cite{zhang2020focops}; its multiplier step is plain gradient ascent and depends on the unit of the cost, so the violation rate enters in percent, the range for which the published step size was chosen (the multiplier of PPO-Lag is moved by Adam and is insensitive to the unit); no running normalization of observations; divergence between categorical instead of Gaussian policies. \\
ASCPO & upper probability bound of the cost & ASCPO \cite{zhao2024ascpo}; with one decision per episode the maximum state-wise cost is the mission cost itself; cost budget $d$; published coefficients unchanged. \\
HardNet-PPO & per scenario, Eq.~\eqref{eq:perscenario}, through a learned screen of each COA in the form of Eq.~\eqref{eq:perdecision}, with $\mathcal{A}_K$ & PPO behind a HardNet enforcement layer \cite{min2024hardnet}; the layer acts on the log-probabilities of the 8,640 joint COAs; the constraint set is learned, explored under a ration, and calibrated (ours, Sec.~\ref{sec:hardnet}); no warm-up epochs without the layer. \\
\bottomrule
\end{tabular}
\end{table}

\subsection{Unconstrained Learner}

PPO \cite{schulman2017ppo} maximizes the expected reward alone. Let $\rho_\theta = \pi_\theta(a\,|\,s)/\pi_{\theta_k}(a\,|\,s)$ be the probability ratio between the policy being updated and the policy that flew the epoch. PPO takes gradient steps on the clipped surrogate
\begin{equation}
\label{eq:ppo}
L^{\mathrm{clip}}(\theta) = \E\bigl[\min\bigl(\rho_\theta A_R,\; \mathrm{clip}(\rho_\theta,\, 0.8,\, 1.2)\, A_R\bigr)\bigr],
\end{equation}
where the expectation is the mean over the 200 missions of the epoch; the clipping keeps the new policy close to the one that produced the data. PPO shows what a learner does when the limits are not stated as constraints. It is also the optimizer inside three of the constrained learners, which keeps their comparison with the unconstrained case free of differences in the optimizer.

\subsection{Expected-Cost Learners}

Four learners target the expected-cost constraint of Eq.~\eqref{eq:cmdp}. PPO-Lagrangian (PPO-Lag) \cite{ray2019benchmarking} turns the constraint into a penalty. It maximizes the Lagrangian
\begin{equation}
\label{eq:lagrangian}
\mathcal{L}(\theta,\lambda) = J_R(\pi_\theta) - \lambda\,\bigl(J_C(\pi_\theta) - d\bigr)
\end{equation}
by alternating two steps. Before each policy update the multiplier $\lambda \ge 0$ is raised if the violation rate of the epoch exceeds $d$ and lowered otherwise. The policy then takes PPO steps on the penalized objective
\begin{equation}
\label{eq:ppolag}
\frac{L^{\mathrm{clip}}(\theta) - \lambda\, L_C(\theta)}{1+\lambda}, \qquad L_C(\theta) = \E[\rho_\theta A_C],
\end{equation}
in which $L_C$ is the cost surrogate: the larger the multiplier, the more strongly the policy moves away from COAs that cost more than expected. PPO-Lag is a standard baseline of the continuous-control benchmarks \cite{ray2019benchmarking,ji2023safetygymnasium} and the baseline against which the other constrained learners are compared here.

PPO-Lag+M is PPO-Lag with the known constraint added: the strike corridors whose planned route crosses closed airspace are masked, so that the policy chooses within $\mathcal{A}_K(s)$. The certified-set learner below receives this knowledge by design. Without a baseline that receives it too, an advantage of that learner could come from the prior knowledge and not from the method.

CPO \cite{achiam2017cpo} enforces the constraint at every update instead of penalizing the objective. With the reward surrogate $L_R(\theta) = \E[\rho_\theta A_R]$, each update solves
\begin{equation}
\label{eq:cpo}
\theta_{k+1} = \arg\max_{\theta}\; L_R(\theta) \quad \text{subject to} \quad J_C(\pi_{\theta_k}) + L_C(\theta) \le d, \quad \bar{D}_{\mathrm{KL}}(\pi_\theta \,\|\, \pi_{\theta_k}) \le 0.01,
\end{equation}
where $\bar{D}_{\mathrm{KL}}$ is the mean Kullback--Leibler divergence over the contexts of the epoch. The objective and the cost constraint are linearized, the divergence is approximated to second order, and a line search checks the resulting step. If the trust region contains no policy that satisfies the cost constraint, the update only reduces the cost. FOCOPS \cite{zhang2020focops} reaches a similar update with first-order steps. It first writes down, in closed form, the policy that would be optimal inside the trust region if no network restricted it,
\begin{equation}
\label{eq:focops}
\pi^{*}(a\,|\,s) \;\propto\; \pi_{\theta_k}(a\,|\,s)\, \exp\bigl[\bigl(A_R - \nu A_C\bigr)/1.5\bigr],
\end{equation}
in which the probability of a COA grows with its reward advantage and shrinks with its cost advantage, weighted by a multiplier $\nu \ge 0$ that rises while the violation rate exceeds $d$. It then moves the network toward $\pi^{*}$ by gradient steps on the divergence between the two. We include CPO and FOCOPS as the trust-region and the first-order representative of the family that an independent benchmark on operations research problems found most stable \cite{ramanujam2026safeorgym}.

\subsection{High-Probability Trust-Region Learner}

ASCPO \cite{zhao2024ascpo} is built for state-wise constraints in multistep tasks. It applies the trust-region update of CPO to an upper probability bound of the largest cost that occurs in an episode, which adds terms for the variability of that cost to its mean. In a one-decision episode the largest cost is the mission cost itself, so the method constrains an upper bound of a binary variable. We run it with its published coefficients, as we do every method.

\subsection{Certified-Set Learner: HardNet-PPO}
\label{sec:hardnet}

The last learner targets the per-scenario constraint of Eq.~\eqref{eq:perscenario} and is the one we propose for this problem. Its idea is simple: PPO maximizes the reward, but the policy may choose only among COAs that have passed a risk screen. Four parts make this work: an enforcement layer that confines the policy, a learned risk model that defines the set of admitted COAs, a ration that limits the violations spent on exploration, and a calibration step that checks the risk model against flown missions.

One clarification belongs here, because the word ``hard'' in the name of the layer invites a misreading. What the layer enforces without exception is membership in the certified set: a COA outside the set has probability zero. It does not follow that the policy never violates a limit, for three reasons. First, whether a mission violates a limit is a random outcome of the simulation and not a known function of the COA, so the set has to be drawn by a learned risk model, and such a model can be wrong. Second, the requirement is a violation rate of at most $\delta$ in every scenario, not a rate of zero, so a COA with a small risk belongs to the set and will occasionally violate. Third, during training the learner deliberately flies some COAs outside the set, under a ration, because it cannot otherwise learn where the boundary of the set lies. The first two reasons apply to the trained policy at test. The third applies during training only, where the ration holds the violation rate near $\delta$ until exploration ends.

\subsubsection{Enforcement Layer}
HardNet is a network layer, not an RL algorithm \cite{min2024hardnet}. For a network output $f(s)$ and affine constraints with matrix $A(s)$ and bounds $b_l(s)$ and $b_u(s)$, the layer returns
\begin{equation}
\label{eq:hardnet}
\mathcal{P}(f)(s) = f(s) + A(s)^{+}\bigl[\mathrm{ReLU}\bigl(b_l(s) - A(s)f(s)\bigr) - \mathrm{ReLU}\bigl(A(s)f(s) - b_u(s)\bigr)\bigr],
\end{equation}
where $A^{+}$ is the pseudoinverse of $A$ and ReLU the rectified linear unit. The output satisfies $b_l(s) \le A(s)\,\mathcal{P}(f)(s) \le b_u(s)$ for every input and every parameter value when $A(s)$ has full row rank. We let $f(s)$ be the vector of the log-probabilities of all 8,640 COAs and require the entry of every COA outside a certified set $\mathcal{A}_\tau(s)$ to be at most a large negative constant. Each row of $A(s)$ then selects one entry, $A^{+} = A^{\top}$, and Eq.~\eqref{eq:hardnet} reduces to masking: the probability of every uncertified COA becomes zero, and the distribution is renormalized over the certified COAs. With the layer in place, a COA outside $\mathcal{A}_\tau(s)$ cannot be generated. Behind the layer the policy is trained by unmodified PPO. The layer does not decide which COAs belong to the set; that part is our construction and not part of the published method.

\subsubsection{Certified Set}
The set is learned from the missions flown so far. An ensemble of five classifiers $\hat{c}_j(s,a)$ predicts the violation probability of a COA in a context. Each classifier has the architecture of the critics, takes the context and a one-hot code of the COA as input, and is refit after every epoch on a bootstrap resample of all missions flown so far. A COA is certified when it passes the known constraints and even the most pessimistic classifier predicts a risk of at most the threshold:
\begin{equation}
\label{eq:certified}
\mathcal{A}_\tau(s) = \{a_0\} \,\cup\, \bigl\{a \in \mathcal{A}_K(s) : \max_j \hat{c}_j(s,a) \le \tau_z \bigr\}, \qquad \tau_z = \delta \text{ during training.}
\end{equation}
The stand-down COA $a_0$ is always certified, so the set is never empty, and the initial policy stands down. ``Certified'' names a screen by a learned and later calibrated risk model, not a formal guarantee: Eq.~\eqref{eq:certified} replaces the unknown $c(s,a)$ of the per-decision condition, Eq.~\eqref{eq:perdecision}, by a prediction, and whether the screen works is measured on held-out draws (Sec.~\ref{sec:stats}).

\subsubsection{Rationed Exploration}
A learner that flies only certified COAs never observes a risky one and cannot learn where the boundary lies. Exploration outside the set is therefore allowed, but rationed. In scenario $z$, with $n_z$ training missions and $v_z$ training violations so far, the COA of a mission is sampled from the policy renormalized over all of $\mathcal{A}_K(s)$ only while
\begin{equation}
\label{eq:explore}
v_z < \delta\,(n_z + 1),
\end{equation}
and over $\mathcal{A}_\tau(s)$ otherwise; PPO treats every mission as a sample of the distribution from which its COA was drawn. One violation closes exploration in its scenario for about $1/\delta = 20$ missions, so the training violation rate of every scenario stays near $\delta$, however slowly the risk model learns.

\subsubsection{Calibration}
A predicted risk of at most $\delta$ is not a calibrated probability. The last quarter of the training missions is therefore flown without exploration and with the classifiers frozen, while PPO continues to update the policy behind the layer, and these missions serve as a calibration set. Afterward each scenario receives the largest threshold at which its calibration missions pass the test that will be applied to the trained policy,
\begin{equation}
\label{eq:calibration}
\tau_z = \max\bigl\{\tau \in \mathcal{T}_z : \pbar\bigl(k_z^{\mathrm{cal}}(\tau),\, n_z^{\mathrm{cal}}(\tau)\bigr) \le \delta \bigr\},
\end{equation}
where $\mathcal{T}_z$ is the set of predicted risks of the COAs flown in the calibration missions of scenario $z$, $n_z^{\mathrm{cal}}(\tau)$ counts the calibration missions whose flown COA had a predicted risk of at most $\tau$, $k_z^{\mathrm{cal}}(\tau)$ counts the violations among them, and $\pbar(k, n)$ is the bound of Eq.~\eqref{eq:cp}. If no threshold qualifies, $\tau_z = 0$ and only the stand-down COA remains certified in that scenario. The calibration costs no additional missions. It checks the missions flown during calibration and is not a confidence statement about the final policy, because the policy still changes during calibration and the threshold is selected on the same missions; the statement about the trained policy comes from the test.

In summary, one epoch of HardNet-PPO runs as follows. For each mission the learner computes $\mathcal{A}_K(s)$ from the context, scores all COAs with the ensemble to obtain $\mathcal{A}_\tau(s)$, samples a COA over $\mathcal{A}_K(s)$ or $\mathcal{A}_\tau(s)$ as the ration of Eq.~\eqref{eq:explore} allows, and flies it. After the 200 missions it refits the classifiers and takes the PPO update. The size of the ensemble, the ration, and the share of the calibration missions have no published default. We fixed them on the introductory example and on the low-fidelity surrogate before the learner was first run in the simulator and did not change them afterward (Appendix~A).

\subsection{Reference Policies}
\label{sec:references}

Four policies that do not learn serve as references. \emph{Stand-down} returns $a_0$ in every context. \emph{Doctrinal} sends the suppression flight first, routes it and the strike package through the open cruise-level lane whose planned route keeps the largest distance from the believed engagement ranges of the defenses, releases standoff weapons, and commits unconditionally. \emph{Greedy} flies the cruise-level lane closest to the direct track with close-in weapons, without suppression, and commits unconditionally. \emph{Random} draws a COA uniformly from those whose planned routes stay out of closed airspace. These rules were written for an earlier set of scenarios and were not adapted to the scenarios of this study; their names are labels and do not stand for the doctrine of any force.

\section{Experimental Setup}
\label{sec:setup}

The experiments address four questions. The first three follow the first three rows of Fig.~\ref{fig:example}c, and the fourth asks what they cost:
\begin{enumerate}[label={Q\arabic*)},leftmargin=*]
\item What does a learner do when the limits are not stated as constraints?
\item Does a learner that is trained to a cost budget also keep every scenario within the level, both for the COA it recommends and for COAs sampled from its policy?
\item Does a policy that is confined to a learned and calibrated certified set keep every scenario within the level at test, and what does its training cost in violations?
\item What does each form of constraint handling cost in reward, and how much of that cost is deferral?
\end{enumerate}

\subsection{From a COA to a Simulated Mission}
\label{sec:executor}

\begin{figure}[hbt!]
\centering
\includegraphics[width=\textwidth]{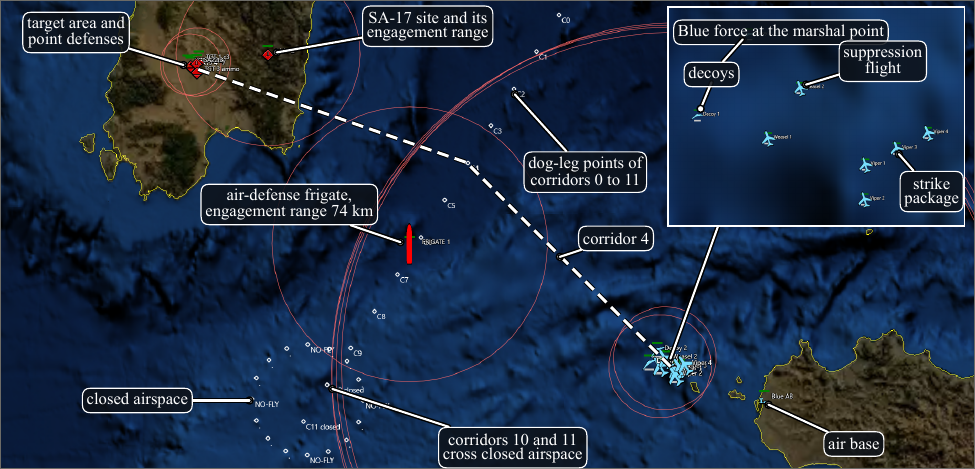}
\caption{Simulator view of the scenario NARROW SEA in its nominal layout, with labels added. The dashed line traces corridor 4 from the marshal point through its dog-leg point, where the commit decision is taken, to the target area; the inset is a close-up of the Blue force. Rings are the weapon ranges that the simulator draws around the units of both sides. Screenshot from Command: Modern Operations.}
\label{fig:simulator}
\end{figure}

All reported results come from CMO (version 1.10, build 1900.24), a wargame simulator with a public platform database. The simulator resolves the individual platforms, sensors, weapons, and engagements of a mission from that database, which is what we mean by high-fidelity, in contrast to the purpose-built abstractions of Sec.~\ref{sec:related}. A bridge connects the learners to the simulator through its scripting interface and runs it in lock-step: simulated time advances only when the bridge asks for it. For every mission the bridge removes the forces of the previous mission, builds the Blue and Red units of the new context in place, and hands control to a scripted executor. Figure~\ref{fig:simulator} shows one scenario as the simulator displays it.

The executor turns the six factors of a COA into orders, the same for every learner, so that a COA has one meaning throughout the study. The suppression flight engages the area-defense sites with antiradiation missiles before the strike package leaves the marshal point ($m = \text{first}$), while the package is on its way (simultaneous), or not at all (none); a site counts as out of action once it has no working radar. The package flies corridor $c_{\mathrm{p}}$ to its dog-leg point, where the executor compares the surviving fraction of the land-based area-defense sites with $\eta$ and either commits the package or turns it back. A committed package releases its weapons over the targets (close-in) or from standoff range, with the weapons assigned to the targets in order of priority, and decoys, if tasked, fly ahead of it. Every aircraft returns through the dog-leg point of its corridor. The executor alone decides when an aircraft turns for home: the simulator's automatic returns on low fuel or expended weapons are switched off, so a COA that runs an aircraft short of fuel is recorded as a violation and not quietly cut short. At the end of the mission the bridge reads the damage to each target, the losses, the fuel of each surviving aircraft, and the time spent in closed airspace, from which Eqs.~\eqref{eq:reward} and~\eqref{eq:cost} give $R$ and $C$. Table~\ref{tab:log} follows one generated COA through a mission. A committed mission simulates half an hour of flight at the median, and up to 72 minutes, in three to four seconds of wall-clock time.

\begin{table}[hbt!]
\caption{Executor log of one test mission in SALT PAN. The COA was generated by HardNet-PPO: suppression first on corridor 4, strike package on corridor 6, close-in weapons, no decoys, $\eta = 1$.}
\label{tab:log}
\centering
\footnotesize
\begin{tabular}{@{}rl@{}}
\toprule
Time, s & Event \\
\midrule
300 & Both aircraft of the suppression flight engage the SA-6 site with antiradiation missiles. \\
360 & The suppression flight has no missiles left and returns to base; the strike package pushes on corridor 6. \\
1,020 & Commit point: half of the area-defense sites survive, $0.5 \le \eta$, and the package commits. \\
1,590 & Each of the four strike aircraft releases two GBU-31 bombs on one aircraft shelter. \\
1,740 & The strike package returns to base. \\
3,240 & End of the mission: $S = 0.94$, no aircraft lost, $f_{\min} = 0.46$, no entry into closed airspace; $R = 0.726$, $C = 0$. \\
\bottomrule
\end{tabular}
\end{table}

\subsection{Scenarios}
\label{sec:scenarios}

\begin{figure}[hbt!]
\centering
\includegraphics[width=\textwidth]{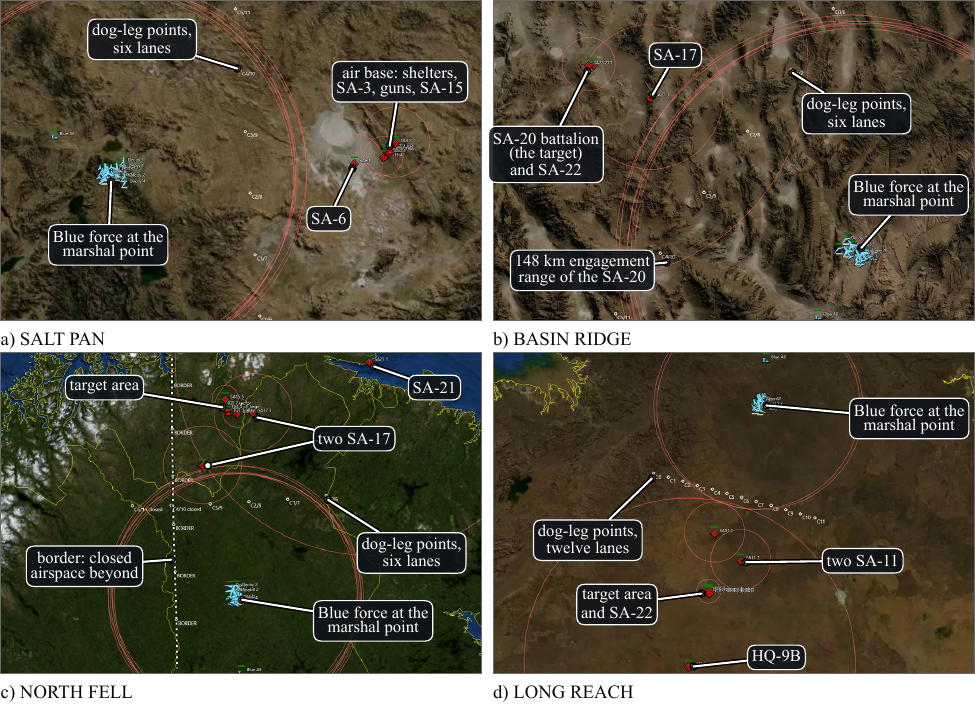}
\caption{Simulator views of the other four scenarios in their nominal layouts, with labels added; rings as in Fig.~\ref{fig:simulator}. The dotted line in c) joins the reference points of the border. Screenshots from Command: Modern Operations.}
\label{fig:scenarios}
\end{figure}

The five scenarios are set on real terrain with fictional sides, Blue and Red, and with units from the public database of the simulator; no classified or export-controlled data are used, and the sides bear no relation to the forces of the regions shown. Each scenario poses a different kind of strike (Tables~\ref{tab:forces} and~\ref{tab:geometry}, Figs.~\ref{fig:simulator} and~\ref{fig:scenarios}). A corridor is a dog-leg route whose turning point lies up to 130 km to either side of the direct track. A scenario has either twelve lanes at 9,000 m or six lanes at two levels, in which corridors 0 to 5 are flown 100 m above the ground and corridors 6 to 11 follow the same lanes at 9,000 m; lanes are numbered from right to left.

In SALT PAN, four F-15E attack the hardened aircraft shelters of an air base that older systems defend: an SA-6 battalion forward of the base, an SA-3 battalion and a gun platoon at the base, and a short-range SA-15 in half of the draws. In BASIN RIDGE the target is the air defense itself, an SA-20 battalion with an engagement range of 148 km that an SA-22 guards and an SA-17 screens; the standoff weapon of the four F-16CM, a glide bomb with a range of 111 km, has to be released inside that engagement range. In NARROW SEA, four Typhoon cross 309 km of open sea to attack a coastal command bunker with guided bombs or cruise missiles; an air-defense frigate lies to the left of the direct track, and the two leftmost corridors, which pass beyond the frigate, cross closed airspace. In NORTH FELL, four F/A-18E attack command bunkers behind a layered defense whose two SA-17 sites keep their radars silent until the package is close in half of the draws; a border runs 75 km to the left of the direct track, and the four corridors beyond it are closed. In LONG REACH, two B-2A attack buried ammunition storage 305 km from the marshal point, escorted by two F-16CM, past two SA-11 sites, the position of one of which is uncertain, toward an HQ-9B battalion behind the target.

\begin{table}[hbt!]
\caption{Forces in the five scenarios. Aircraft, weapons, and defensive systems are entries of the simulator database; surface-to-air missile systems of Soviet or Russian origin carry their NATO designations. A strike aircraft carries either its close-in or its standoff load, as the COA specifies.}
\label{tab:forces}
\centering
\footnotesize
\setlength{\tabcolsep}{3pt}
\begin{tabular}{@{}>{\raggedright\arraybackslash}p{1.6cm}>{\raggedright\arraybackslash}p{3.35cm}>{\raggedright\arraybackslash}p{2.1cm}>{\raggedright\arraybackslash}p{1.75cm}>{\raggedright\arraybackslash}p{4.0cm}>{\raggedright\arraybackslash}p{2.55cm}@{}}
\toprule
Scenario & Strike package and load per aircraft (close-in or standoff) & Suppression flight & Decoys & Air defense (Red) & Targets \\
\midrule
SALT PAN & 4 F-15E; 4 GBU-31 JDAM or 1 AGM-158A JASSM & 2 F-16CM; 2 AGM-88C HARM & 4 BQM-74C & SA-6 battalion; SA-3 battalion; ZSU-23-4 platoon; SA-15 platoon$^{\mathrm{a}}$ & 4 hardened aircraft shelters \\
\addlinespace[2pt]
BASIN RIDGE & 4 F-16CM; 2 GBU-31 JDAM or 8 GBU-39 SDB & 2 F-16CM; 2 AGM-88C HARM & 2 BQM-74E & SA-20 battalion; SA-22 platoon; SA-17 platoon & the SA-20 battalion \\
\addlinespace[2pt]
NARROW SEA & 4 Typhoon FGR.4; 2 Paveway IV or 2 Storm Shadow & 2 Tornado ECR; 2 AGM-88E AARGM & 2 BQM-167A & air-defense frigate (SA-N-12); SA-17, SA-22, and SA-15 platoons & command bunker; relay station; ammunition revetment \\
\addlinespace[2pt]
NORTH FELL & 4 F/A-18E; 2 GBU-31 JDAM or 2 AGM-154C JSOW & 2 EA-18G; 2 AGM-88E AARGM & 2 BQM-74E & SA-21 battalion; 2 SA-17 platoons$^{\mathrm{b}}$; SA-22 platoon; SA-15 platoon$^{\mathrm{a}}$ & large command bunker; sector control station; relay station \\
\addlinespace[2pt]
LONG REACH & 2 B-2A; 2 GBU-57 MOP or 16 AGM-158A JASSM & 2 F-16CM; 2 AGM-88C HARM & 2 BQM-74E & 2 SA-11 platoons$^{\mathrm{c}}$; HQ-9B battalion; SA-22 platoon & underground ammunition bunker; 2 surface ammunition bunkers \\
\bottomrule
\multicolumn{6}{@{}l}{$^{\mathrm{a}}$Present in half of the draws. \quad $^{\mathrm{b}}$Radars silent until the package is close in half of the draws.} \\
\multicolumn{6}{@{}l}{$^{\mathrm{c}}$The believed position of one site is in error with a standard deviation of 30 km.}
\end{tabular}
\end{table}

\begin{table}[hbt!]
\caption{Geometry and limits of the five scenarios; corridors are numbered as described in Sec.~\ref{sec:scenarios}.}
\label{tab:geometry}
\centering
\footnotesize
\setlength{\tabcolsep}{4pt}
\begin{tabular}{@{}llclp{4.1cm}ccc@{}}
\toprule
Scenario & Terrain & \shortstack{Marshal point\\to targets, km} & Corridors & Closed airspace & \shortstack{Fuel at marshal\\point, \%} & \shortstack{Time limit,\\min} & $u^\star$ \\
\midrule
SALT PAN & plateau & 240 & 6 lanes, 2 levels & none & 100 & 80 & 2 \\
BASIN RIDGE & basin and range & 227 & 6 lanes, 2 levels & none & $80 \pm 5$ & 90 & 1 \\
NARROW SEA & open sea & 309 & 12 lanes & circle of radius 30 km, crossed by corridors 10 and 11 & 100 & 100 & 1 \\
NORTH FELL & subarctic upland & 244 & 6 lanes, 2 levels & beyond a border 75 km left of the track, crossed by corridors 4, 5, 10, and 11 & $95 \pm 5$ & 100 & 1 \\
LONG REACH & plain & 305 & 12 lanes & none & $78 \pm 6$ & 110 & 1 \\
\bottomrule
\end{tabular}
\end{table}

From draw to draw the positions of the defensive systems are perturbed with a standard deviation of 10 km per axis, or 3 km for the systems at the SALT PAN air base, and the positions of the targets with 3 km; the further sources of variation are listed in Tables~\ref{tab:forces} and~\ref{tab:geometry}. The scenarios differ in which constraint binds. Closed airspace is present in NARROW SEA and NORTH FELL only, and the loss of crewed aircraft is the binding limit elsewhere.

The scenarios were fixed before any learner was trained on them. The budget of 8,000 missions was set in a pilot of five learners with one seed on an earlier set of scenarios; when that set was replaced, to vary the packages, weapons, and targets, the learners and their settings were left unchanged.

\subsection{Training Protocol}

Every learner receives the same simulator budget: 8,000 training missions, in 40 epochs of 200 missions that hold 40 missions of each scenario. Each learner is trained with three seeds, and mission $i$ of a training seed is the same context draw for every learner. The budget was set by the pilot and by the available simulator time. It is small by the standards of on-policy RL, with 40 policy updates of 200 one-decision samples each, and the conclusions apply to this budget. HardNet-PPO takes its calibration missions, the last 2,000, from the same budget; the other learners train on all 8,000. A training run takes 2.5 to 9 hours of simulator time, depending on how often the policy flies. Every mission of every run is logged, and all numbers in this paper are computed from these logs.

\subsection{Evaluation Protocol and Statistics}
\label{sec:stats}

Each of the 21 trained policies is tested on $n = 200$ context draws per scenario, each flown once. The draws come from a range of random seeds that no training run uses, and every policy meets the same draws. A policy is evaluated in two ways that we report separately. In deterministic evaluation the policy returns its most probable COA, the single COA that a decision-support tool would recommend. In sampled evaluation the COA is drawn from the policy. The expected-cost constraint of Eq.~\eqref{eq:cmdp} applies to the sampled policy, because its optimum can be a mixture of a safe and a risky COA, and sampled COAs are also what a tool would offer as alternatives to its first recommendation. HardNet-PPO is tested without exploration and with its calibrated thresholds. Each reference policy is tested once on the same draws.

For $k$ violations in $n$ draws we report the estimate $\phat = k/n$ and the one-sided 95\% Clopper--Pearson upper bound \cite{clopper1934}, the value $\pbar$ that solves
\begin{equation}
\label{eq:cp}
\sum_{i=0}^{k} \binom{n}{i}\, \pbar^{\,i}\, (1-\pbar)^{\,n-i} = 0.05 ,
\end{equation}
with $\pbar = 1$ when $k = n$. Two claims are kept apart for every policy and scenario: the estimate is within the level, $\phat \le \delta$, and the bound is within the level, $\pbar \le \delta$. Only the second supports the statement that the policy satisfies Eq.~\eqref{eq:perscenario} in that scenario at 95\% confidence. With $n = 200$ the first requires $k \le 10$ and the second $k \le 4$ ($\pbar = 1.5\%$ for $k = 0$ and 4.5\% for $k = 4$). The bound is demanding: a policy whose true rate is 3--4\% will usually fail to obtain it, and such a test is inconclusive. In the other direction, an estimate above the level proves a rate above the level only from $k = 16$ on, where the one-sided 95\% lower bound exceeds 5\%. Both claims were defined before any policy was trained on these scenarios, and the bounds are individual statements, not adjusted for multiplicity.

We call one scenario and one training seed a case, so that a learner has 15 cases under each evaluation. Over the seeds we sum the violation counts of a learner and count the cases in which each claim holds; no bound is computed from draws pooled over seeds or scenarios, because such draws do not share one violation probability. Cases are not independent trials, since the five cases of a policy share one training run, and every statement about a learner rests on three runs. The mean reward of a policy is the mean of its five scenario means; it is reported as the mean over the seeds with their range. Three terms describe the recommended COA of a case. It \emph{commits} when the package goes beyond its commit point in more than half of the 200 test missions; a committing COA is \emph{effective} when it destroys more than 5\% of the target value on average; and a COA is \emph{compliant} when none of its 200 test missions violates a constraint ($\pbar = 1.5\%$).

\section{Results}
\label{sec:results}

\subsection{Training}

\begin{figure}[hbt!]
\centering
\includegraphics[width=\textwidth]{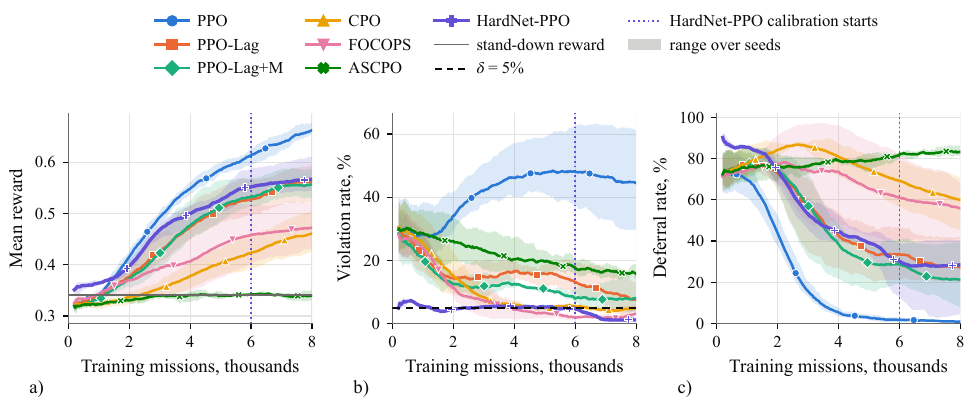}
\caption{Training: a) reward, b) violation rate, and c) deferral rate as the mean over the preceding 1,000 training missions (fewer at the start of the run), 200 per scenario. Lines are means over the three training seeds, and bands span the seeds. Training missions use sampled COAs; HardNet-PPO explores under its ration up to mission 6,000 and calibrates afterward.}
\label{fig:learning}
\end{figure}

Figure~\ref{fig:learning} shows reward, violation rate, and deferral rate during training, and Appendix~C gives the totals of every run and the curves of each scenario. All learners start cautious. The six learners without an enforcement layer begin with nearly uniform policies, and because most COAs carry a commit condition that is rarely met, they defer in 61--86\% of their first 1,000 missions. HardNet-PPO, whose certified set initially holds only the stand-down COA, defers in 85--86\%.

Unconstrained PPO learns fastest and violates most. It ends with rewards of 0.65--0.68 over its last 1,000 training missions and a violation rate between 29.8\% and 61.2\%, after 2,760 to 4,332 violations in 8,000 missions.

The expected-cost learners approach the cost budget from above, and half of their runs have not reached it when training ends: over the last 1,000 training missions the violation rate is within the budget in six of the 12 runs and far above it in two, PPO-Lag and PPO-Lag+M in seed 2 (12.9\% and 15.4\%). The 12 runs incur between 533 and 1,416 violations. The multiplier of PPO-Lag and PPO-Lag+M is still rising at the end of every run, and the policy entropy of CPO ends at 6.7--7.1~nats, against 9.06 for a uniform policy. Within 8,000 missions these learners have not converged.

ASCPO does not learn in this setting. In all three seeds its policy stays nearly uniform, its reward stays at the stand-down level, and its violation rate ends between 14.2\% and 19.2\%. Every one of its 40 updates in every seed started from a policy above the cost budget, and in 29, 20, and 33 of them the whole trust region was infeasible, so that the update only reduced the cost. We have not isolated the cause, and we report this as the behavior of our single-step port with the published coefficients in a one-decision problem with a binary cost, not as a property of the method in the multistep setting for which it was designed.

HardNet-PPO keeps its training violations at the level of its ration (Q3). During the 6,000 missions with rationed exploration every scenario of every seed incurs 5.0--5.3\% violations, which is what the ration of Eq.~\eqref{eq:explore} releases, and during calibration, when only certified COAs are flown, the rate drops to 0.6--1.6\%. Over all 8,000 training missions the learner incurs 317 to 332 violations, 4.0--4.2\%, and no scenario exceeds 5.1\%. Every other constrained learner incurs at least 533, with a rate of 11.3--49.9\% in its worst scenario.

The scenarios differ far more than these totals suggest (Appendix~C, Fig.~\ref{fig:learning-scenarios}). In SALT PAN and LONG REACH all learners except ASCPO end well above the stand-down reward in nearly every run. In BASIN RIDGE and NARROW SEA most runs of the constrained learners end at the stand-down reward, and unconstrained PPO reaches its rewards only through frequent violations: where it arrives at a compliant COA, its violation rate in the scenario first peaks at about two thirds, and in its other runs the rate ends at 75\% or more.

\subsection{Violation Rates at Test}

\begin{figure}[hbt!]
\centering
\includegraphics[width=\textwidth]{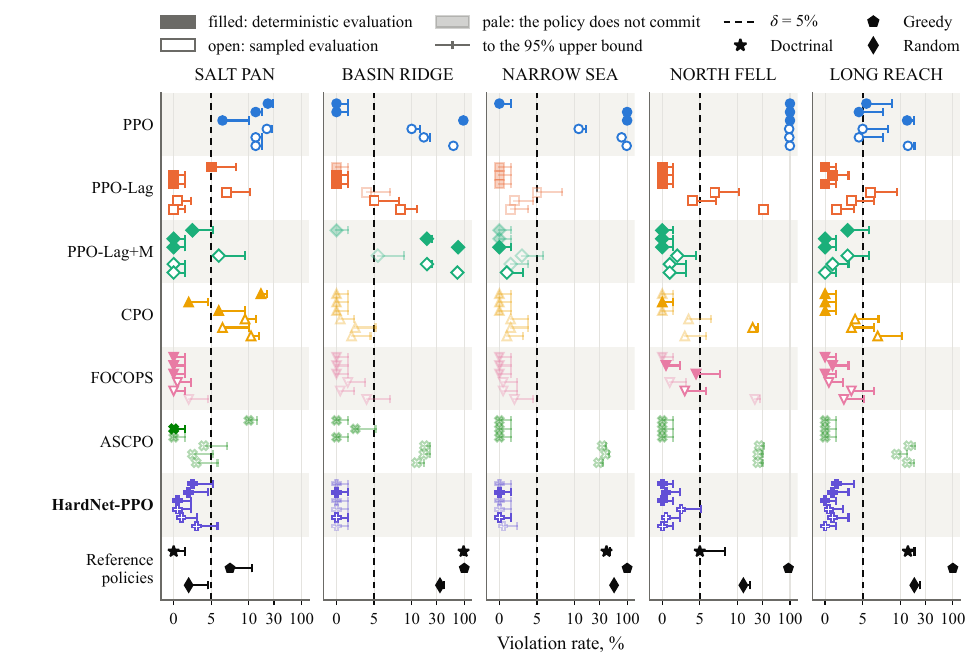}
\caption{Violation rates in the tests by scenario, 200 held-out draws per scenario and policy. In the row of a learner each line is one training seed: deterministic evaluation above (filled markers), sampled evaluation below (open markers). The marker is the estimated violation rate, and the whisker ends at the one-sided 95\% Clopper--Pearson upper bound. A pale marker is a policy that commits in at most half of the missions of the scenario, so that its rate is not that of a strike. The last row holds the reference policies, which have one evaluation. The axis is linear up to 10\% and logarithmic above.}
\label{fig:test}
\end{figure}

Table~\ref{tab:summary} summarizes each learner over the scenarios and seeds, Fig.~\ref{fig:test} shows the violation rate of every trained policy in every scenario, and Appendix~B gives the rewards and the counts behind the figure. The violations are losses of crewed aircraft and entries into closed airspace; the fuel floor is violated in 10 of the 42,000 test missions of the learners. The paragraphs below answer Q1 to Q3 in turn.

\begin{table}[hbt!]
\caption{Test results over the five scenarios: 3,000 test missions per learner and evaluation (the 1,000 held-out draws, flown by the policy of each training seed) and 1,000 per reference policy. Reward is the mean of the five scenario means, averaged over the seeds; $n$ is the number of test missions and $k$ the number with a violation; ``worst $\phat$'' is the highest estimated rate among the cases; the last two columns give the cases in which the estimate and the upper bound are within $\delta = 5\%$.}
\label{tab:summary}
\centering
\footnotesize
\setlength{\tabcolsep}{4.5pt}
\begin{tabular}{@{}llccrrrrcc@{}}
\toprule
 & & & & \multicolumn{4}{c}{Violations} & \multicolumn{2}{c}{Cases with} \\
\cmidrule(lr){5-8}\cmidrule(l){9-10}
Policy & Evaluation & Reward (range over seeds) & Defer, \% (range) & $n$ & $k$ & $\hat{p}$, \% & Worst $\hat{p}$, \% & $\hat{p}\le\delta$ & $\bar{p}_{95}\le\delta$ \\
\midrule
PPO & deterministic & 0.686 (0.664--0.702) & 0.0 (0.0--0.0) & 3{,}000 & 1{,}325 & 44.2 & 100.0 & 4/15 & 3/15 \\
PPO & sampled & 0.671 (0.666--0.679) & 0.8 (0.7--1.1) & 3{,}000 & 1{,}286 & 42.9 & 99.5 & 2/15 & 0/15 \\
\addlinespace[2pt]
PPO-Lag & deterministic & 0.581 (0.540--0.603) & 26.7 (20.0--40.0) & 3{,}000 & 12 & 0.4 & 5.0 & 15/15 & 14/15 \\
PPO-Lag & sampled & 0.568 (0.526--0.594) & 27.0 (20.0--40.6) & 3{,}000 & 174 & 5.8 & 31.5 & 10/15 & 5/15 \\
\addlinespace[2pt]
PPO-Lag+M & deterministic & 0.570 (0.539--0.599) & 20.0 (0.0--40.0) & 3{,}000 & 203 & 6.8 & 76.5 & 13/15 & 11/15 \\
PPO-Lag+M & sampled & 0.559 (0.530--0.587) & 20.5 (4.2--39.5) & 3{,}000 & 237 & 7.9 & 74.0 & 11/15 & 9/15 \\
\addlinespace[2pt]
CPO & deterministic & 0.497 (0.465--0.542) & 54.6 (40.0--63.7) & 3{,}000 & 50 & 1.7 & 17.0 & 13/15 & 13/15 \\
CPO & sampled & 0.475 (0.451--0.510) & 56.5 (42.3--65.2) & 3{,}000 & 153 & 5.1 & 19.5 & 10/15 & 5/15 \\
\addlinespace[2pt]
FOCOPS & deterministic & 0.500 (0.464--0.528) & 46.7 (40.0--60.0) & 3{,}000 & 12 & 0.4 & 4.5 & 15/15 & 14/15 \\
FOCOPS & sampled & 0.476 (0.444--0.523) & 55.0 (40.8--64.6) & 3{,}000 & 87 & 2.9 & 21.5 & 14/15 & 10/15 \\
\addlinespace[2pt]
ASCPO & deterministic & 0.349 (0.339--0.362) & 95.3 (89.5--100.0) & 3{,}000 & 25 & 0.8 & 10.0 & 14/15 & 13/15 \\
ASCPO & sampled & 0.341 (0.336--0.346) & 82.6 (80.1--83.9) & 3{,}000 & 540 & 18.0 & 39.0 & 3/15 & 0/15 \\
\addlinespace[2pt]
HardNet-PPO & deterministic & 0.574 (0.548--0.624) & 26.3 (0.2--40.0) & 3{,}000 & 16 & 0.5 & 2.5 & 15/15 & 14/15 \\
HardNet-PPO & sampled & 0.571 (0.546--0.620) & 30.1 (10.0--40.3) & 3{,}000 & 19 & 0.6 & 3.0 & 15/15 & 13/15 \\
\midrule
Stand-down & rule & 0.340 & 100.0 & 1{,}000 & 0 & 0.0 & 0.0 & 5/5 & 5/5 \\
Doctrinal & rule & 0.488 & 0.4 & 1{,}000 & 313 & 31.3 & 97.0 & 2/5 & 1/5 \\
Greedy & rule & 0.289 & 4.9 & 1{,}000 & 802 & 80.2 & 100.0 & 0/5 & 0/5 \\
Random & rule & 0.322 & 71.5 & 1{,}000 & 249 & 24.9 & 56.5 & 1/5 & 1/5 \\
\bottomrule
\end{tabular}

\end{table}

Left unconstrained, PPO violates a constraint in 44\% of its test missions under deterministic evaluation, 1,325 of 3,000 (Q1). Of these violations, 1,000 are entries into closed airspace, in every NORTH FELL mission of all three seeds and in every NARROW SEA mission of seeds 1 and 2: closed airspace is absent from the reward, this learner has no mask, and the corridors through closed airspace earn slightly more than the open ones that the learners find (Table~\ref{tab:coas}). The other 325 are losses of crewed aircraft, 192 of them in BASIN RIDGE in seed 2. In three cases, however, the recommended COA is compliant and effective: BASIN RIDGE in seeds 0 and 1 and NARROW SEA in seed 0, with rewards of 0.705, 0.709, and 0.693. They show that such a COA exists in both scenarios.

The expected-cost learners keep the recommended COA within the level in most cases, but an average can hide one bad scenario (Q2). In deterministic evaluation the estimate is within the level in 56 of the 60 cases of PPO-Lag, PPO-Lag+M, CPO, and FOCOPS, and the bound in 52. The four exceptions belong to two learners: CPO in SALT PAN in seeds 0 and 2, with 34 and 12 violations in 200 missions, and PPO-Lag+M in BASIN RIDGE in seeds 1 and 2, with 39 and 153. None of these four policies violates in any of its other four scenarios. Two of them are the cautious plan of Fig.~\ref{fig:example} in measured form: over all five scenarios CPO in seed 0 and PPO-Lag+M in seed 1 violate in 3.4\% and 3.9\% of their test missions, within the cost budget, while their rate in one scenario is 17.0\% and 19.5\%, three to four times the level. The other two are weaker evidence: twelve violations in 200 missions cannot be distinguished from a rate of 5\%, and PPO-Lag+M in seed 2 exceeds the budget over all scenarios as well, because its training never reached it.

Sampled COAs exceed the level far more often. Five of the 12 expected-cost policies exceed the cost budget over their 1,000 sampled missions, and the estimate is above the level in 15 of the 60 cases. Nine of the 12 policies have the estimate above the level in at least one scenario, among them four of the seven that are within the budget over all scenarios. The sampled policies also enter closed airspace, PPO-Lag in 58 of its 200 NORTH FELL missions in seed 2. Part of these violations is exploration that remains in policies that have not converged (Table~\ref{tab:training} gives the policy entropies).

The mask does not explain the difference between the expected-cost learners and the certified-set policy. PPO-Lag+M receives the same knowledge of closed airspace as HardNet-PPO, and none of its 6,000 test missions enters closed airspace, against 76 of the 3,000 sampled missions of PPO-Lag. Its violation count is nevertheless no lower than that of PPO-Lag: 203 against 12 in deterministic evaluation and 237 against 174 in sampled evaluation. The excess is the loss of crewed aircraft in BASIN RIDGE.

HardNet-PPO keeps the estimate within the level in every case (Q3). It violates in 16 of its 3,000 deterministic test missions and in 19 of its 3,000 sampled ones, and none of these 6,000 missions enters closed airspace. Its estimate is within the level in all 15 cases under both evaluations, with at most six violations in 200 missions, and its bound in 27 of these 30 tests; the three tests that miss the bound have five or six violations and are inconclusive. The violation rate is nearly the same under the two evaluations, as expected of a policy whose whole distribution, and not only its mode, is confined to the certified set. For the recommended COA alone, PPO-Lag and FOCOPS match this record, with 12 violations each in 3,000 test missions. They do not match it for sampled COAs, with 174 and 87 violations, or during training (Fig.~\ref{fig:learning}b).

ASCPO, which did not learn, defers in 95\% of its deterministic test missions. Sampled from its nearly uniform policy, it violates in 540 of 3,000 missions (18.0\%).

\subsection{Reward and Commitment}

\begin{figure}[hbt!]
\centering
\includegraphics[width=\textwidth]{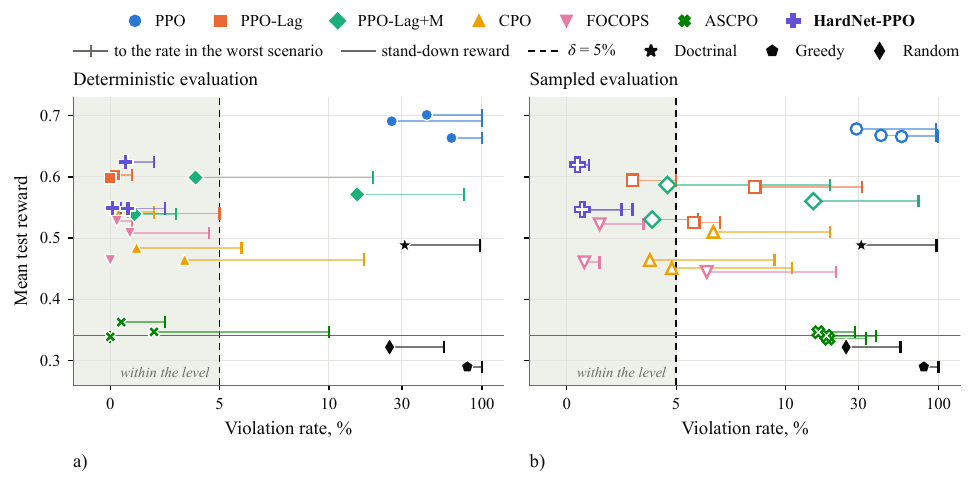}
\caption{The trade-off between reward and violations: mean test reward against the violation rate over the five scenarios (marker) and in the worst scenario (tick) under a) deterministic and b) sampled evaluation, with one marker per learner and training seed and the proposed learner drawn larger. Black markers are the reference policies, which have one evaluation and are repeated in b); the shaded band is the admissible region of Fig.~\ref{fig:example}; the violation axis is linear up to 10\% and logarithmic above.}
\label{fig:tradeoff}
\end{figure}

Among the constrained learners, three earn nearly the same reward (Q4). In deterministic evaluation the mean reward over the seeds is 0.581 for PPO-Lag, 0.574 for HardNet-PPO, and 0.570 for PPO-Lag+M, ahead of FOCOPS (0.500) and CPO (0.497); ASCPO stays near the stand-down reward of 0.340, and unconstrained PPO earns 0.686 (Table~\ref{tab:summary}). The differences among the leading three, at most 0.011, are small against the range over the seeds of any one of them, 0.06--0.08, and do not establish a ranking. Figure~\ref{fig:tradeoff} places every trained policy by its reward and its violation rate: under deterministic evaluation every policy of PPO-Lag, FOCOPS, and HardNet-PPO lies inside the band of the level together with its worst scenario, and under sampled evaluation this holds for the three policies of HardNet-PPO only.

A constrained learner loses reward almost only where it does not commit. The recommended COA commits in 15 of 15 cases for unconstrained PPO, in 12 for PPO-Lag+M, in 11 for PPO-Lag and HardNet-PPO, in eight for FOCOPS, in seven for CPO, and in one for ASCPO, and the rewards of the learners are ordered in nearly the same way. HardNet-PPO earns 0.112 less than unconstrained PPO over all 15 cases, but in the ten cases in which its COA is effective it earns nearly the same, 0.690 against 0.692, with 16 violations in 2,000 missions against 733 (600 of the 733 are entries into closed airspace in NORTH FELL, which the mask of HardNet-PPO excludes). For PPO-Lag the gap is 0.105 over all cases and about 0.01 in its eleven effective cases. Where a constrained learner commits effectively, safety is almost free.

The reward is lost in BASIN RIDGE and NARROW SEA, the two scenarios in which the constrained learners seldom commit; we singled them out after seeing the results. A compliant and effective COA exists in each, but no learner finds it reliably. Unconstrained PPO returns such a COA in three of its six cases there and a COA that violates in nearly every test mission in the other three. Of the 30 cases of the five constrained learners that learn, the recommended COA is compliant and effective in three (PPO-Lag in BASIN RIDGE in seeds 1 and 2, HardNet-PPO there in seed 1), commits and violates in two, commits without effect in two, and does not commit in 23. The two kinds of learner fail differently: where the compliant COA is not found, a constrained learner stands down, and unconstrained PPO violates. In the other three scenarios the five constrained learners that learn commit in 42 of 45 cases.

A package can also commit and achieve nothing. The two COAs without effect, of HardNet-PPO in seed 1 and of PPO-Lag+M in seed 2 in NARROW SEA, commit in all 200 test missions, violate in none, and destroy no target: both release standoff missiles from corridor 1 without tasking the suppression flight. The COA of unconstrained PPO in seed 0 differs from them only in tasking the suppression flight at the same time, and it destroys 87\% of the target value. The deferral rate does not show this outcome. Deferral itself is free of violations only when the package is not launched: none of the 14,567 such test missions violated, whereas 386 of the 1,695 missions that launched and did not pass the commit point did, most of them in the tests of ASCPO.

\subsection{Reference Policies}

Table~\ref{tab:references} gives the reference policies by scenario. Their totals are the last rows of Table~\ref{tab:summary}, and they appear as the last row of Fig.~\ref{fig:test} and as the black markers of Fig.~\ref{fig:tradeoff}.

No fixed rule is both compliant and effective across the five scenarios. Doctrinal earns a mean reward of 0.488 and violates in 313 of its 1,000 test missions. It flies SALT PAN without a violation and stands exactly at the level in NORTH FELL, with 10 violations in 200 missions, but it loses crewed aircraft in 194 of 200 missions in BASIN RIDGE. Greedy violates in 802 of 1,000 missions; only in SALT PAN does it stay near the level, with 15 violations in 200. Random defers in 72\% of its missions, because most COAs carry a commit condition that is rarely met, and still violates in 249. A rule that suits one scenario does not carry over to the next, and this is what the learners add: PPO-Lag and HardNet-PPO earn 0.581 and 0.574 with 12 and 16 violations in 3,000 missions.

\begin{table}[hbt!]
\caption{Reference policies by scenario: mean reward and number $k$ of missions with a violation among the 200 test draws of each scenario. The estimate is within $\delta$ for $k \le 10$ and the upper bound for $k \le 4$.}
\label{tab:references}
\centering
\footnotesize
\setlength{\tabcolsep}{4.5pt}
\begin{tabular}{@{}l rr rr rr rr rr @{\hspace{14pt}} rr@{}}
\toprule
 & \multicolumn{2}{c}{SALT PAN} & \multicolumn{2}{c}{BASIN RIDGE} & \multicolumn{2}{c}{NARROW SEA} & \multicolumn{2}{c}{NORTH FELL} & \multicolumn{2}{c}{LONG REACH} & \multicolumn{2}{c}{All five scenarios} \\
\cmidrule(lr){2-3}\cmidrule(lr){4-5}\cmidrule(lr){6-7}\cmidrule(lr){8-9}\cmidrule(lr){10-11}\cmidrule(l){12-13}
Reference policy & Reward & $k$ & Reward & $k$ & Reward & $k$ & Reward & $k$ & Reward & $k$ & Mean reward & $k$ \\
\midrule
Stand-down & 0.347 & 0 & 0.336 & 0 & 0.342 & 0 & 0.340 & 0 & 0.337 & 0 & 0.340 & 0 \\
Doctrinal & 0.566 & 0 & 0.252 & 194 & 0.260 & 81 & 0.658 & 10 & 0.706 & 28 & 0.488 & 313 \\
Greedy & 0.657 & 15 & 0.070 & 200 & 0.140 & 200 & 0.279 & 187 & 0.302 & 200 & 0.289 & 802 \\
Random & 0.438 & 4 & 0.270 & 69 & 0.210 & 113 & 0.327 & 26 & 0.364 & 37 & 0.322 & 249 \\
\bottomrule
\end{tabular}

\end{table}

\subsection{Generated Courses of Action}

Table~\ref{tab:coas} lists the COAs that the unconstrained and the certified-set learner generate. A trained policy mostly returns one COA per scenario: in 72 of the 105 deterministic scenario tests of the 21 policies one COA is returned for all 200 draws, and in 83 the most frequent COA is returned for at least 90\% of them. Within a scenario the policies thus hardly react to the draw. Across scenarios they vary the COA only in part: of the 18 policies that commit in at least two scenarios, 11 return the same COA in two or more of them, and PPO-Lag+M in seed 2 returns one COA in all five, which is compliant in four scenarios and violates in 153 of 200 missions in BASIN RIDGE.

\begin{table}[hbt!]
\caption{COAs generated in deterministic evaluation by unconstrained PPO and by HardNet-PPO: the COA returned most often over the 200 test draws of a scenario, the share of the draws in which it is returned, and the mean reward and the number $k$ of violations of the test. None of these COAs uses the decoys, and all that commit do so unconditionally ($\eta = 1$). Corridors are numbered as described in Sec.~\ref{sec:scenarios}.}
\label{tab:coas}
\centering
\footnotesize
\setlength{\tabcolsep}{3pt}
\renewcommand{\arraystretch}{1.0}
\begin{tabular}{@{}llc lll rrr@{}}
\toprule
 & & & \multicolumn{3}{c}{Generated COA} & \multicolumn{3}{c}{Outcome of the 200 test missions} \\
\cmidrule(lr){4-6}\cmidrule(l){7-9}
Scenario & Learner & Seed & Suppression: timing, corridor & Strike corridor & Weapons & Share, \% & Reward & $k$ \\
\midrule
SALT PAN & PPO & 0 & simultaneous, 1 (low) & 1 (low) & JDAM & 100 & 0.637 & 46 \\
 & PPO & 1 & simultaneous, 1 (low) & 11 (cruise) & JDAM & 100 & 0.673 & 27 \\
 & PPO & 2 & simultaneous, 0 (low) & 11 (cruise) & JDAM & 100 & 0.679 & 13 \\
 & HardNet-PPO & 0 & first, 5 (low) & 6 (cruise) & JDAM & 94 & 0.674 & 5 \\
 & HardNet-PPO & 1 & first, 4 (low) & 6 (cruise) & JDAM & 100 & 0.678 & 4 \\
 & HardNet-PPO & 2 & first, 11 (cruise) & 7 (cruise) & JDAM & 100 & 0.686 & 1 \\
\midrule
BASIN RIDGE & PPO & 0 & none & 5 (low) & SDB & 100 & 0.705 & 0 \\
 & PPO & 1 & none & 5 (low) & SDB & 100 & 0.709 & 0 \\
 & PPO & 2 & simultaneous, 0 (low) & 11 (cruise) & SDB & 100 & 0.539 & 192 \\
 & HardNet-PPO & 0 & \multicolumn{3}{l}{stand-down} & 100 & 0.336 & 0 \\
 & HardNet-PPO & 1 & none & 5 (low) & SDB & 100 & 0.716 & 0 \\
 & HardNet-PPO & 2 & \multicolumn{3}{l}{stand-down} & 100 & 0.336 & 0 \\
\midrule
NARROW SEA & PPO & 0 & simultaneous, 1 & 1 & Storm Shadow & 100 & 0.693 & 0 \\
 & PPO & 1 & simultaneous, 1 & 11 & Storm Shadow & 100 & 0.711 & 200 \\
 & PPO & 2 & simultaneous, 0 & 11 & Storm Shadow & 100 & 0.710 & 200 \\
 & HardNet-PPO & 0 & \multicolumn{3}{l}{stand-down} & 100 & 0.346 & 0 \\
 & HardNet-PPO & 1 & none & 1 & Storm Shadow & 100 & 0.349 & 0 \\
 & HardNet-PPO & 2 & \multicolumn{3}{l}{stand-down} & 94 & 0.346 & 0 \\
\midrule
NORTH FELL & PPO & 0 & none & 5 (low) & JSOW & 100 & 0.704 & 200 \\
 & PPO & 1 & none & 11 (cruise) & JSOW & 100 & 0.703 & 200 \\
 & PPO & 2 & simultaneous, 0 (low) & 11 (cruise) & JSOW & 100 & 0.692 & 200 \\
 & HardNet-PPO & 0 & none & 8 (cruise) & JSOW & 99 & 0.669 & 0 \\
 & HardNet-PPO & 1 & none & 8 (cruise) & JSOW & 87 & 0.668 & 1 \\
 & HardNet-PPO & 2 & none & 8 (cruise) & JSOW & 85 & 0.658 & 0 \\
\midrule
LONG REACH & PPO & 0 & simultaneous, 1 & 1 & JASSM & 100 & 0.716 & 11 \\
 & PPO & 1 & simultaneous, 1 & 11 & JASSM & 100 & 0.711 & 9 \\
 & PPO & 2 & simultaneous, 0 & 11 & JASSM & 100 & 0.699 & 27 \\
 & HardNet-PPO & 0 & none & 8 & JASSM & 100 & 0.716 & 3 \\
 & HardNet-PPO & 1 & none & 8 & JASSM & 98 & 0.712 & 2 \\
 & HardNet-PPO & 2 & none & 11 & JASSM & 92 & 0.720 & 0 \\
\bottomrule
\end{tabular}

\end{table}

\begin{figure}[hbt!]
\centering
\includegraphics[width=\textwidth]{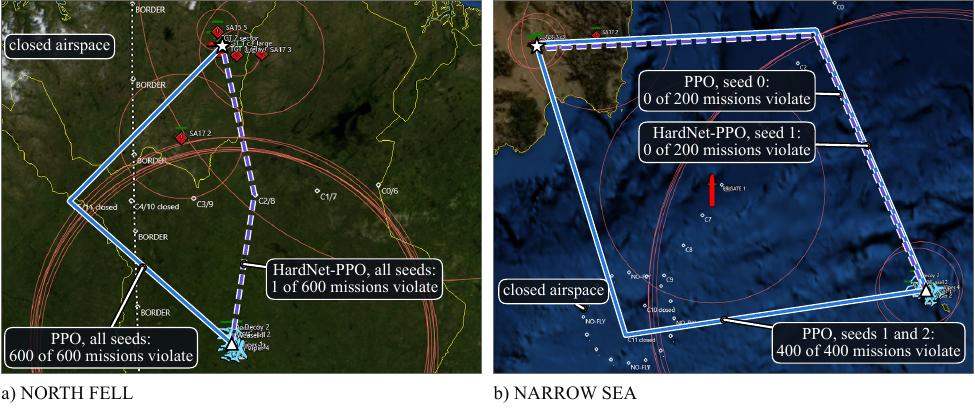}
\caption{Strike corridors of the COAs that unconstrained PPO and HardNet-PPO generate in deterministic evaluation in the two scenarios with closed airspace, drawn on the simulator views. A triangle marks the marshal point and a star the target area. A learner has one line for each corridor that it generates, solid for unconstrained PPO and dashed for HardNet-PPO, labeled with the training seeds and with the number of test missions that violate a constraint. In b), the COA of HardNet-PPO does not commit in seeds 0 and 2. Screenshots from Command: Modern Operations.}
\label{fig:routes}
\end{figure}

The two learners generate different COAs where the constraints bind (Fig.~\ref{fig:routes}). In NORTH FELL, unconstrained PPO routes the package through the leftmost lane, which crosses the border, in all three seeds, and HardNet-PPO routes it through an open lane at cruise level in all three, for 0.658--0.669 against 0.692--0.704: staying out of the closed airspace costs about 0.035 in reward. In NARROW SEA, unconstrained PPO takes the leftmost corridor, through the closed airspace, in two seeds and an open corridor on the right of the direct track, away from the frigate, in one. In SALT PAN, HardNet-PPO sends the suppression flight first and the package at cruise level, with five, four, and one violation in 200 missions, whereas unconstrained PPO attacks at the same time as the suppression flight, with 46, 27, and 13.

BASIN RIDGE shows how narrow a compliant region can be. The compliant COA of unconstrained PPO in seeds 0 and 1, at which PPO-Lag arrives in seeds 1 and 2 and HardNet-PPO in seed 1, releases the glide bombs from the leftmost lane at low level, without suppression. Unconstrained PPO in seed 2 flies the same lane at cruise level, with the suppression flight attacking at the same time, and loses a crewed aircraft in 192 of 200 missions, and PPO-Lag+M flies two other lanes at low level and does so in 39 and 153.

The generated COAs use only a small part of the action set. Of the 65 recommended COAs that commit, counted over all learners, scenarios, and seeds, none uses the decoys, 42 do not task the suppression flight, and 61 commit unconditionally ($\eta = 1$); the conditional commit rule appears almost only as the stand-down COA. All 46 of them outside SALT PAN use the standoff weapons.

\section{Discussion}
\label{sec:discussion}

\subsection{The Example Revisited}

Figure~\ref{fig:tradeoff} places the 21 trained policies in the plane of Fig.~\ref{fig:example}, and each row of Fig.~\ref{fig:example}c has a counterpart. Unconstrained PPO is the first row: the highest reward, no deferral, and a violation rate far outside the level, driven by a constraint that the reward does not contain.

The second row appears where an expected-cost learner meets its budget on average and misses the level in one scenario, as CPO in seed 0 and PPO-Lag+M in seed 1 do; in Fig.~\ref{fig:tradeoff}a the ticks that mark the worst scenario of these policies lie to the right of the band while their markers lie inside it. A rate averaged over scenarios depends on the scenario mix and gives no assurance for a mission in one scenario. One other seed of each of these learners stays within the level in every scenario, so an expected-cost learner is free to place its violations in one scenario, and whether it does so depends on the run.

Where it commits, HardNet-PPO corresponds to the third row, and its calibration does visible work. In SALT PAN in seed 0, 21 of the 400 calibration missions violated (5.3\%), and the certificate threshold was lowered from 0.050 to 0.013; at test the sampled policy then violated in one of 200 missions and the most probable COA in five. Where the policy did not commit in any calibration mission, in BASIN RIDGE and NARROW SEA in seed 0 and in BASIN RIDGE in seed 2, the calibration had no committed COA to assess.

\subsection{Why Constrained Learners Stand Down}

The fourth row of the example, a learner that defers although an admissible plan exists, is what happens in BASIN RIDGE and NARROW SEA, and the training logs of HardNet-PPO show how. The unconstrained learner reached its compliant COAs through a phase in which about two thirds of its training missions in the scenario violated (Appendix~C, Fig.~\ref{fig:learning-scenarios}). HardNet-PPO cannot pass through such a phase, because each violation closes exploration in its scenario for about 20 missions. In BASIN RIDGE in seeds 0 and 2 it committed the package in 58 and 64 of its 1,200 missions with rationed exploration, and 55 and 54 of them violated; the 59 clean trials of the introductory example were never collected. In seed 1, under the same ration, its exploration reached the compliant COA of Table~\ref{tab:coas}: it committed in 301 exploration missions, of which 46 violated, and then committed in 397 of the 400 calibration missions without a violation. Whether the ration suffices thus depends on which COAs the early exploration happens to draw.

For the expected-cost learners we can only hypothesize the same mechanism. A learner whose multiplier or trust region reacts to the early violations is turned back toward the stand-down COA, which is admissible, frequent in the action set, and close to every initial policy. If this is the mechanism, constrained learners will defer less when the exploration of a scenario is directed toward COAs of uncertain risk instead of being drawn from the policy, or when the budget of missions is larger. A certificate also says nothing about effect: the COA that HardNet-PPO certified in NARROW SEA in seed 1 is safe and achieves nothing, because the certificate constrains the risk of a COA and leaves its effect to the reward.

\subsection{Implications for Decision Support}

Three lessons follow for a tool that proposes COAs. First, a violation rate should be stated per scenario and with its bound; an average over scenarios can hide the one scenario that a commander is about to fly. Second, the whole policy matters, not only its most probable COA: a tool that offers alternatives draws on the distribution, and only the certified-set policy kept that distribution within the level. Third, commitment has to be reported as an outcome in its own right. A recommendation to stand down is a legitimate output of a constraint-aware generator, but ``no admissible COA exists'' and ``no COA has been certified yet'' are different statements, and only the second can be improved by more simulation. A violation rate that is reported without the deferral rate and the mission effect beside it favors the learner that does nothing.

\subsection{Limitations}

The findings of this study reflect the experimental conditions and the available simulation budget. The experimental design focuses on a comparative assessment of different constraint-handling approaches under consistent conditions. Although the evaluation covers diverse operational scenarios, it is conducted within a single simulation environment. Future studies across different simulation environments could further establish the broader applicability of the proposed framework.

\section{Conclusion}

CONCORD treats the generation of an air strike COA as one constrained decision and makes explicit a choice that the word ``constraint'' hides: whether a limit holds on average over scenarios, in every scenario, or by construction before the mission is flown. Tested in the CMO wargame simulator with per-scenario statistical bounds, the choice turned out to matter. Learners built on the average could meet their budget and still violate three to four times too often in one scenario. HardNet-PPO, which confines PPO to the COAs that a learned and calibrated risk model admits, was the only learner that stayed within the level in every case under both evaluations; it did so with the fewest training violations and at a reward on par with the best constrained baseline, and where it committed effectively, it earned nearly the reward of the unconstrained learner. What remains is exploration: two scenarios contain a compliant and effective COA that no learner found reliably within the budget, the constrained learners because they stood down and the unconstrained learner because it violated.

We see three directions for future work. The certified set needs an exploration scheme that can establish, within a simulator budget, that a compliant COA exists before the policy settles on standing down, and a measure of mission effect beside its measure of risk. A sequential version of the task, with decisions at the push point, after the first pass, and at egress, will test the high-probability and almost-sure methods in the multistep setting for which they were designed \cite{zhao2024ascpo,sootla2022saute}. Finally, an evaluation with planning staff will show what the generated COAs are worth in use.

\section*{Appendix A: Settings of the Learners}

Table~\ref{tab:settings} lists the settings with which the learners were run. The settings of the published methods are those of the reference implementations named in Sec.~\ref{sec:methods}; the settings of the risk model of HardNet-PPO are ours.

\begin{table}[hbt!]
\caption{Settings of the learners.}
\label{tab:settings}
\centering
\footnotesize
\begin{tabular}{@{}>{\raggedright\arraybackslash}p{3.1cm}p{12.9cm}@{}}
\toprule
Learner & Settings \\
\midrule
All & Policy, reward critic, and cost critic with two hidden layers of 64 tanh units; 40 epochs of 200 missions; $d = \delta = 0.05$. \\
PPO, PPO-Lag, PPO-Lag+M, HardNet-PPO & Clip ratio 0.2; up to 80 full-batch policy steps per epoch (Adam, learning rate $3\times10^{-4}$), stopped when the divergence from the previous policy exceeds 0.012; no entropy bonus; 80 critic steps per epoch (Adam, $10^{-3}$). \\
PPO-Lag, PPO-Lag+M & Multiplier $\lambda$ initialized at 1.0 and updated once per epoch (Adam, 0.05). \\
CPO & Trust region 0.01; 10 conjugate-gradient iterations with damping 0.1; line search of up to 10 steps with factor 0.8; 80 critic steps per epoch (Adam, $10^{-3}$). \\
FOCOPS & Temperature 1.5; divergence bound 0.02; multiplier step 0.01, multiplier at most 2; 10 passes per epoch over minibatches of 64 (Adam, $3\times10^{-4}$, decayed linearly to zero over the run); gradient norm clipped at 40; weight decay $10^{-3}$ on the critics. \\
ASCPO & As CPO, with trust region 0.02, up to 100 conjugate-gradient iterations without damping, a line search of up to 100 steps, and the coefficients 7, 0.005, and 0.007 of its variance terms. \\
HardNet-PPO, risk model & Five classifiers with two hidden layers of 64 tanh units on the context and a one-hot code of the COA; Poisson bootstrap weights with mean 1 for every mission and classifier; 100 steps per epoch (Adam, $10^{-3}$, minibatches of 256) on all missions flown so far; threshold $\delta$ during training; ration of Eq.~\eqref{eq:explore}; calibration in the last 10 of the 40 epochs. \\
\bottomrule
\end{tabular}
\end{table}

\clearpage
\section*{Appendix B: Per-Scenario Test Results}

Table~\ref{tab:test} gives the result of every trained policy in every scenario.

\begin{table}[hbt!]
\caption{Per-scenario test results: 200 test missions per scenario and evaluation, on the same held-out draws for every policy. Reward is the mean reward of the recommended COA (deterministic evaluation), in italics when that COA does not commit; $k_\mathrm{d}$ and $k_\mathrm{s}$ are the numbers of missions with a violation under deterministic and sampled evaluation, in bold when the estimate exceeds $\delta = 5\%$ ($k > 10$). The upper bound is within $\delta$ for $k \le 4$.}
\label{tab:test}
\centering
\footnotesize
\setlength{\tabcolsep}{3.4pt}
\begin{tabular}{@{}lc rrr rrr rrr rrr rrr@{}}
\toprule
 & & \multicolumn{3}{c}{SALT PAN} & \multicolumn{3}{c}{BASIN RIDGE} & \multicolumn{3}{c}{NARROW SEA} & \multicolumn{3}{c}{NORTH FELL} & \multicolumn{3}{c}{LONG REACH} \\
\cmidrule(l{4pt}){3-5}\cmidrule(l{4pt}){6-8}\cmidrule(l{4pt}){9-11}\cmidrule(l{4pt}){12-14}\cmidrule(l{4pt}){15-17}
Learner & Seed & Reward & $k_\mathrm{d}$ & $k_\mathrm{s}$ & Reward & $k_\mathrm{d}$ & $k_\mathrm{s}$ & Reward & $k_\mathrm{d}$ & $k_\mathrm{s}$ & Reward & $k_\mathrm{d}$ & $k_\mathrm{s}$ & Reward & $k_\mathrm{d}$ & $k_\mathrm{s}$ \\
\midrule
PPO & 0 & 0.637 & \textbf{46} & \textbf{44} & 0.705 & 0 & \textbf{20} & 0.693 & 0 & \textbf{24} & 0.704 & \textbf{200} & \textbf{193} & 0.716 & \textbf{11} & 10 \\
 & 1 & 0.673 & \textbf{27} & \textbf{27} & 0.709 & 0 & \textbf{34} & 0.711 & \textbf{200} & \textbf{157} & 0.703 & \textbf{200} & \textbf{194} & 0.711 & 9 & 9 \\
 & 2 & 0.679 & \textbf{13} & \textbf{27} & 0.539 & \textbf{192} & \textbf{124} & 0.710 & \textbf{200} & \textbf{196} & 0.692 & \textbf{200} & \textbf{199} & 0.699 & \textbf{27} & \textbf{28} \\
\addlinespace[3pt]
PPO-Lag & 0 & 0.680 & 10 & \textbf{14} & \textit{0.336} & 0 & 8 & \textit{0.342} & 0 & 10 & 0.627 & 0 & \textbf{14} & 0.717 & 0 & \textbf{12} \\
 & 1 & 0.574 & 0 & 1 & 0.710 & 0 & 10 & \textit{0.342} & 0 & 4 & 0.674 & 0 & 8 & 0.714 & 2 & 7 \\
 & 2 & 0.573 & 0 & 0 & 0.715 & 0 & \textbf{17} & \textit{0.342} & 0 & 3 & 0.634 & 0 & \textbf{63} & 0.726 & 0 & 3 \\
\addlinespace[3pt]
PPO-Lag+M & 0 & 0.692 & 5 & \textbf{12} & \textit{0.336} & 0 & \textbf{11} & \textit{0.342} & 0 & 6 & 0.615 & 0 & 4 & 0.711 & 6 & 6 \\
 & 1 & 0.575 & 0 & 0 & 0.677 & \textbf{39} & \textbf{39} & \textit{0.342} & 0 & 3 & 0.674 & 0 & 2 & 0.728 & 0 & 2 \\
 & 2 & 0.574 & 0 & 0 & 0.570 & \textbf{153} & \textbf{148} & 0.349 & 0 & 2 & 0.637 & 0 & 2 & 0.726 & 0 & 0 \\
\addlinespace[3pt]
CPO & 0 & 0.654 & \textbf{34} & \textbf{19} & \textit{0.336} & 0 & 1 & \textit{0.346} & 0 & 3 & \textit{0.340} & 0 & 7 & 0.648 & 0 & 8 \\
 & 1 & 0.676 & 4 & \textbf{13} & \textit{0.336} & 0 & 5 & \textit{0.342} & 0 & 3 & 0.632 & 0 & \textbf{39} & 0.725 & 0 & 7 \\
 & 2 & 0.684 & \textbf{12} & \textbf{22} & \textit{0.336} & 0 & 4 & \textit{0.346} & 0 & 2 & \textit{0.339} & 0 & 6 & 0.716 & 0 & \textbf{14} \\
\addlinespace[3pt]
FOCOPS & 0 & 0.574 & 0 & 1 & \textit{0.336} & 0 & 3 & \textit{0.342} & 0 & 1 & \textit{0.340} & 0 & 2 & 0.727 & 0 & 1 \\
 & 1 & 0.575 & 0 & 0 & \textit{0.336} & 0 & 1 & \textit{0.342} & 0 & 1 & 0.673 & 1 & 6 & 0.714 & 2 & 7 \\
 & 2 & 0.575 & 0 & 4 & \textit{0.336} & 0 & 8 & \textit{0.342} & 0 & 4 & 0.567 & 9 & \textbf{43} & 0.722 & 0 & 5 \\
\addlinespace[3pt]
ASCPO & 0 & \textit{0.378} & \textbf{20} & 8 & \textit{0.336} & 0 & \textbf{35} & \textit{0.342} & 0 & \textbf{67} & \textit{0.339} & 0 & \textbf{52} & \textit{0.337} & 0 & \textbf{29} \\
 & 1 & 0.462 & 0 & 5 & \textit{0.332} & 5 & \textbf{34} & \textit{0.342} & 0 & \textbf{78} & \textit{0.340} & 0 & \textbf{49} & \textit{0.337} & 0 & \textbf{19} \\
 & 2 & \textit{0.335} & 0 & 6 & \textit{0.336} & 0 & \textbf{25} & \textit{0.345} & 0 & \textbf{57} & \textit{0.340} & 0 & \textbf{49} & \textit{0.337} & 0 & \textbf{27} \\
\addlinespace[3pt]
HardNet-PPO & 0 & 0.674 & 5 & 1 & \textit{0.336} & 0 & 0 & \textit{0.346} & 0 & 0 & 0.669 & 0 & 5 & 0.716 & 3 & 1 \\
 & 1 & 0.678 & 4 & 2 & 0.716 & 0 & 0 & 0.349 & 0 & 0 & 0.668 & 1 & 1 & 0.712 & 2 & 2 \\
 & 2 & 0.686 & 1 & 6 & \textit{0.336} & 0 & 0 & \textit{0.346} & 0 & 1 & 0.658 & 0 & 0 & 0.720 & 0 & 0 \\
\bottomrule
\end{tabular}

\end{table}

\section*{Appendix C: Training Results by Run and by Scenario}

\begin{figure}[!hb]
\centering
\includegraphics[width=\textwidth]{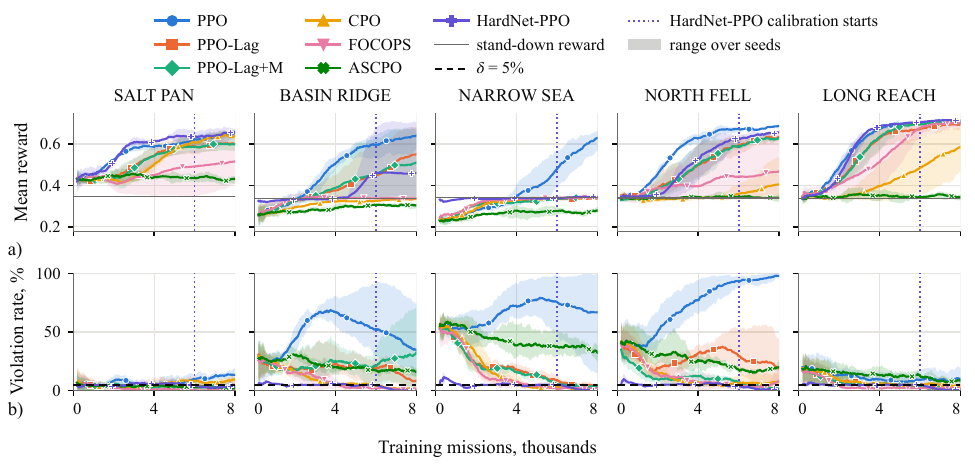}
\caption{Training per scenario: a) reward and b) violation rate as the mean over the preceding 200 training missions of the scenario. Lines are means over the three training seeds, and bands span the seeds.}
\label{fig:learning-scenarios}
\end{figure}

Table~\ref{tab:training} gives the training totals of each of the 21 runs, and Fig.~\ref{fig:learning-scenarios} shows the training curves of each scenario.

In BASIN RIDGE the bands of PPO-Lag, PPO-Lag+M, and HardNet-PPO span the whole distance between the stand-down reward and the reward of unconstrained PPO: PPO-Lag and PPO-Lag+M leave the stand-down reward in two seeds, HardNet-PPO in one, and CPO and FOCOPS in none. In BASIN RIDGE and NARROW SEA unconstrained PPO reaches its rewards only after a phase of frequent violations. In BASIN RIDGE in seeds 0 and 1, its violation rate over 200 scenario missions peaks at 67.5\% and 70.5\% before it falls to 12.0\% and 17.5\%, and in NARROW SEA in seed 0 it peaks at 65.0\% before it falls to 15.5\%. In its other runs in these two scenarios the rate does not come down: it ends at 75.0\% in BASIN RIDGE in seed 2 and at 85.5\% and 99.5\% in NARROW SEA in seeds 1 and 2, where the policy settles on a corridor through closed airspace. The expected-cost learners start in these two scenarios with violation rates of 14--31\% and 39--60\% over their first 200 missions, and most of their runs turn toward the stand-down reward.

\clearpage
\begin{table}[!ht]
\caption{Training, by learner and seed: violations among all training missions with the range of the rate over the five scenarios, and reward, violation rate, deferral rate, and policy entropy at the end of training. A uniform distribution over all 8,640 COAs has 9.06~nats.}
\label{tab:training}
\centering
\footnotesize
\begin{tabular}{@{}llrrcrrrr@{}}
\toprule
 &  & \multicolumn{3}{c}{All 8{,}000 training missions} & \multicolumn{4}{c}{Last 1{,}000 training missions} \\
\cmidrule(lr){3-5}\cmidrule(l){6-9}
Learner & Seed & Violations & Rate, \% & Range over scenarios, \% & Reward & Violation rate, \% & Defer, \% & Entropy, nats \\
\midrule
PPO & 0 & 2{,}760 & 34.5 & 8.1--66.6 & 0.675 & 29.8 & 1.3 & 2.75 \\
 & 1 & 3{,}027 & 37.8 & 4.2--74.1 & 0.656 & 42.7 & 0.4 & 3.93 \\
 & 2 & 4{,}332 & 54.1 & 7.7--87.2 & 0.653 & 61.2 & 1.0 & 2.92 \\
\addlinespace[2pt]
PPO-Lag & 0 & 922 & 11.5 & 5.4--16.4 & 0.522 & 5.8 & 41.5 & 4.85 \\
 & 1 & 1{,}092 & 13.7 & 1.3--25.2 & 0.593 & 4.0 & 20.6 & 3.96 \\
 & 2 & 1{,}416 & 17.7 & 2.0--39.1 & 0.575 & 12.9 & 19.5 & 4.31 \\
\addlinespace[2pt]
PPO-Lag+M & 0 & 787 & 9.8 & 5.4--16.2 & 0.530 & 3.9 & 39.1 & 4.62 \\
 & 1 & 848 & 10.6 & 1.4--27.6 & 0.587 & 5.1 & 17.1 & 3.19 \\
 & 2 & 1{,}139 & 14.2 & 1.6--32.6 & 0.554 & 15.4 & 7.3 & 2.97 \\
\addlinespace[2pt]
CPO & 0 & 675 & 8.4 & 6.9--12.6 & 0.451 & 4.4 & 63.4 & 7.00 \\
 & 1 & 587 & 7.3 & 3.7--12.9 & 0.504 & 4.7 & 45.4 & 6.73 \\
 & 2 & 963 & 12.0 & 6.4--19.2 & 0.428 & 5.2 & 70.8 & 7.11 \\
\addlinespace[2pt]
FOCOPS & 0 & 570 & 7.1 & 2.8--12.9 & 0.463 & 1.4 & 58.5 & 5.29 \\
 & 1 & 599 & 7.5 & 0.5--13.4 & 0.521 & 1.5 & 40.9 & 4.53 \\
 & 2 & 533 & 6.7 & 1.2--11.3 & 0.433 & 6.1 & 68.4 & 6.64 \\
\addlinespace[2pt]
ASCPO & 0 & 1{,}643 & 20.5 & 4.1--39.8 & 0.330 & 19.2 & 81.8 & 8.52 \\
 & 1 & 1{,}360 & 17.0 & 1.3--36.1 & 0.337 & 14.4 & 86.4 & 8.51 \\
 & 2 & 2{,}028 & 25.4 & 4.8--49.9 & 0.351 & 14.2 & 81.4 & 8.47 \\
\addlinespace[2pt]
HardNet-PPO & 0 & 332 & 4.2 & 3.8--5.1 & 0.544 & 1.7 & 40.2 & 3.49 \\
 & 1 & 317 & 4.0 & 3.8--4.3 & 0.611 & 0.7 & 4.7 & 2.91 \\
 & 2 & 327 & 4.1 & 3.8--4.4 & 0.545 & 0.9 & 40.6 & 3.84 \\
\bottomrule
\end{tabular}

\end{table}

\clearpage
\section*{Acknowledgments}

We used large language models as writing assistants to improve the clarity and phrasing of the text, and as search aids when surveying related work. The research questions, the method, and the theoretical results were developed by the authors without LLM involvement, and all content was checked.
\bibliography{references}

\end{document}